\documentclass[10pt,twocolumn,letterpaper]{article}

 \usepackage{cvpr}              

\usepackage{graphicx}
\usepackage{booktabs}

\usepackage[accsupp]{axessibility}  
\usepackage{caption}
\usepackage{bbm}
\usepackage{arydshln}

\usepackage{listings}
\usepackage{placeins}
\usepackage{nicefrac}
\usepackage{xurl}
\usepackage{enumitem}
\usepackage{tabularx}
\usepackage{nicefrac}
\usepackage{makecell}
\usepackage{xspace}
\usepackage{graphbox}
\usepackage{pifont}
\usepackage{enumitem}
\usepackage{colortbl}
\usepackage{multirow}
\usepackage{diagbox}
\usepackage{array}
\usepackage{wrapfig}
\usepackage{subcaption}
\usepackage{footmisc}
\usepackage{color}
\usepackage{stfloats}
\usepackage{slashbox}
\usepackage{diagbox}
\usepackage{float}
\usepackage{mathrsfs}
\usepackage{apptools}
\usepackage{bbm}
\usepackage{soul}

\usepackage[dvipsnames]{xcolor}
\usepackage{tcolorbox}
\usepackage{url}

\definecolor{cvprblue}{rgb}{0.21,0.49,0.74}
\usepackage[pagebackref,breaklinks,colorlinks,citecolor=cvprblue]{hyperref}

\graphicspath{ {figures} }

\def\R{\mathbb{R}}

\newcommand{\inner}[1]{\left\langle#1\right\rangle}

\def\R{\mathbb{R}}

\def\L{\mathcal{L}}

\newcommand{\norm}[1]{\left\|#1\right\|}

\newcommand{\simfn}{\texttt{sim}\xspace}

\newcolumntype{C}[1]{>{\centering\arraybackslash}p{#1}}
\newcolumntype{L}[1]{>{\raggedright\arraybackslash}p{#1}}
\newcolumntype{R}[1]{>{\raggedleft\arraybackslash}p{#1}}
\newlength\newl
\newlength\newlc

\newlength\colwidth
\newlength\figwidth

\newcommand{\oxford}{$\mathcal{R}$\textsc{Oxford}\xspace}
\newcommand{\paris}{$\mathcal{R}$\textsc{Paris}\xspace}

\newcommand{\nights}{\textsc{Nights}\xspace}
\newcommand{\bapps}{\textsc{Bapps}\xspace}
\newcommand{\pieapp}{\textsc{PieAPP}\xspace}
\newcommand{\kadid}{\textsc{Kadid-10k}\xspace}
\newcommand{\koniq}{\textsc{KonIQ-10k}\xspace}
\newcommand{\sice}{\textsc{Sice}\xspace}
\newcommand{\agiqa}{\textsc{Agiqa-3k}\xspace}
\newcommand{\magicbrush}{\textsc{MagicBrush}\xspace}
\newcommand{\imgreward}{\textsc{ImageReward}\xspace}
\newcommand{\hpd}{\textsc{HPDv2}\xspace}
\newcommand{\polaris}{\textsc{Polaris}\xspace}
\newcommand{\hqedit}{\textsc{HQ-Edit}\xspace}
\newcommand{\pipal}{\textsc{Pipal}\xspace}
\newcommand{\imgnet}{\textsc{ImageNet-OOO}\xspace}
\newcommand{\coco}{$\mathcal{CD}$-\textsc{Coco}\xspace}
\newcommand{\mscoco}{\textsc{MS-Coco}\xspace}
\newcommand{\imagenet}{\textsc{ImageNet}\xspace}
\newcommand{\cifarh}{\textsc{Cifar-100}\xspace}
\newcommand{\cifarhooo}{\textsc{Cifar-100-OOO}\xspace}

\newcommand{\gentasks}{\textit{OOD Generalization Tasks}\xspace}
\newcommand{\coretasks}{\textit{Core 2AFC Tasks}\xspace}
\newlength\myindent

\usepackage{amsmath,amsfonts,bm}

\def\eqref#1{Eq.~(\ref{#1})}

\def\1{\bm{1}}

\def\vt{{\bm{t}}}

\def\vx{{\bm{x}}}

\def\vz{{\bm{z}}}

\DeclareMathAlphabet{\mathsfit}{\encodingdefault}{\sfdefault}{m}{sl}
\SetMathAlphabet{\mathsfit}{bold}{\encodingdefault}{\sfdefault}{bx}{n}

\DeclareMathOperator*{\argmax}{arg\,max}

\definecolor{lightgray}{rgb}{0.9,0.9,0.9}
\definecolor{BlueGray}{rgb}{1, 0.8, 0.8}
\definecolor{lightgreen}{rgb}{0.90, 0.99, 0.85}
\definecolor{darkgreen}{rgb}{.1, .85, .1}
\definecolor{newgray}{rgb}{0., 0., 0.} 
\definecolor{graygreen}{rgb}{.1, .75, .1} 
\definecolor{grayred}{rgb}{1., 0., 0.} 
\definecolor{lightorange}{rgb}{1., .9, 0.}
\definecolor{lighttred}{rgb}{1., .9, 0.8}
\definecolor{teaser1}{HTML}{FFCCBC}
\definecolor{teaser1}{HTML}{FFCCBC}

\definecolor{myYellow}{HTML}{FFE599} 
\definecolor{myBlue}{HTML}{CFE2f3} 
\definecolor{myGreen}{HTML}{B6D7A8} 
\definecolor{myRed}{HTML}{EA9999}
\definecolor{myPurple}{HTML}{8E7CC3} 
\definecolor{myPink}{HTML}{C27BA0} 
\definecolor{myOrange}{HTML}{F9CB9C}
\definecolor{myGray}{HTML}{D9D9D9}

\newcommand{\nimgtwoafc}{Img-2AFC\xspace}
\newcommand{\nittwoafc}{IT-2AFC\xspace}
\newcommand{\ntexttwoafc}{Text-2AFC\xspace}
\newcommand{\niqa}{IQA\xspace}
\newcommand{\npaa}{PAA\xspace}
\newcommand{\nooo}{OOO\xspace}
\newcommand{\nir}{IR\xspace}

\newcommand{\imgtoken}{\texttt{<image>}}
\newcommand{\newll}{\texttt{\textbackslash n}}
\newcommand{\prompt}{\texttt{\{prompt\}}}

\newcommand{\imgtwoafc}{\colorbox{myYellow}{Img-2AFC}\xspace}
\newcommand{\ittwoafc}{\colorbox{myBlue}{IT-2AFC}\xspace}
\newcommand{\texttwoafc}{\colorbox{myGreen}{Text-2AFC}\xspace}
\newcommand{\iqa}{\colorbox{myRed}{IQA}\xspace}
\newcommand{\paa}{\colorbox{myPurple}{PAA}\xspace}
\newcommand{\ooo}{\colorbox{myOrange}{OOO}\xspace}
\newcommand{\ir}{\colorbox{myGray}{IR}\xspace}

\newcommand\blfootnote[1]{%
  \begingroup
  \renewcommand\thefootnote{}\footnote{#1}%
  \addtocounter{footnote}{-1}%
  \endgroup
}

\newcommand{\unisimvlm}{
UniSim-CLIP\xspace
}
\newcommand{\unisimlmm}{
UniSim-LL-N\xspace
}
\newcommand{\unisimbnc}{
UniSim-Bench\xspace}

\newcommand{\roundedcellnew}[4]{%
    \begin{tcolorbox}[
        colback=#1, colframe=#1, arc=4pt, boxrule=0pt, width=\linewidth,
        valign=center, height=#2, left=0pt, right=0pt, top=0pt, bottom=0pt, boxsep=0pt, after=\strut
    ]
        \ifstrequal{#3}{left}{\raggedright}{%
            \ifstrequal{#3}{center}{\centering}{%
                \ifstrequal{#3}{right}{\raggedleft}{\centering}}}
        #4
    \end{tcolorbox}
}

\def\paperID{16648} 
\def\confName{CVPR}
\def\confYear{2024}

\title{
Towards Unified Benchmark and Models for Multi-Modal Perceptual Metrics 
}

\author{Sara Ghazanfari$^{1*}$
\qquad Siddharth Garg$^{1}$
\qquad Nicolas Flammarion$^{2}$
\qquad Prashanth Krishnamurthy$^{1}$ \\
\qquad Farshad Khorrami$^{1}$
\qquad Francesco Croce$^{2}$ \vspace{0.6cm} \\ 
$^{1}$New York University, US \qquad $^{2}$EPFL, Switzerland}

\begin{document}

\twocolumn[{
\maketitle
\vspace{-1.0cm}
\begin{center}
    \centering
    
    \includegraphics[width=1.0\textwidth, trim=0mm 10mm 0mm 70mm, clip]{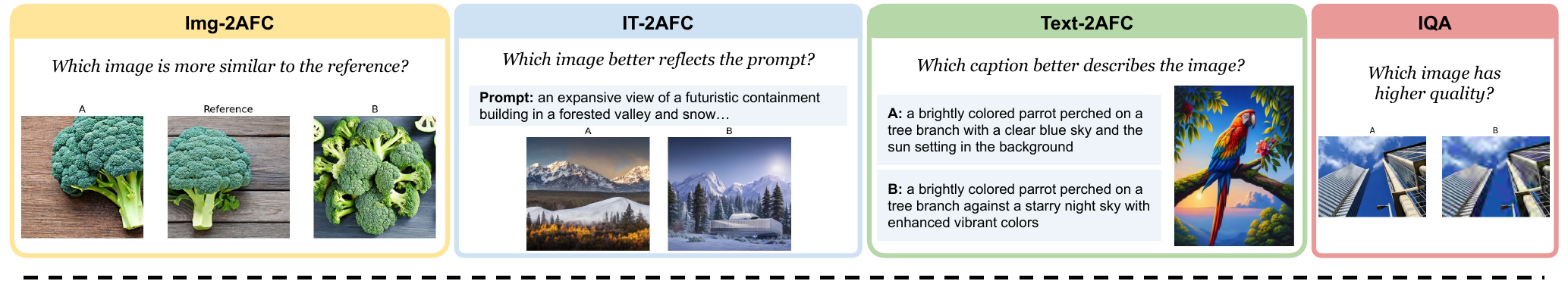}
    \includegraphics[width=.98\textwidth, trim=0mm 0mm 0mm 5mm, clip]{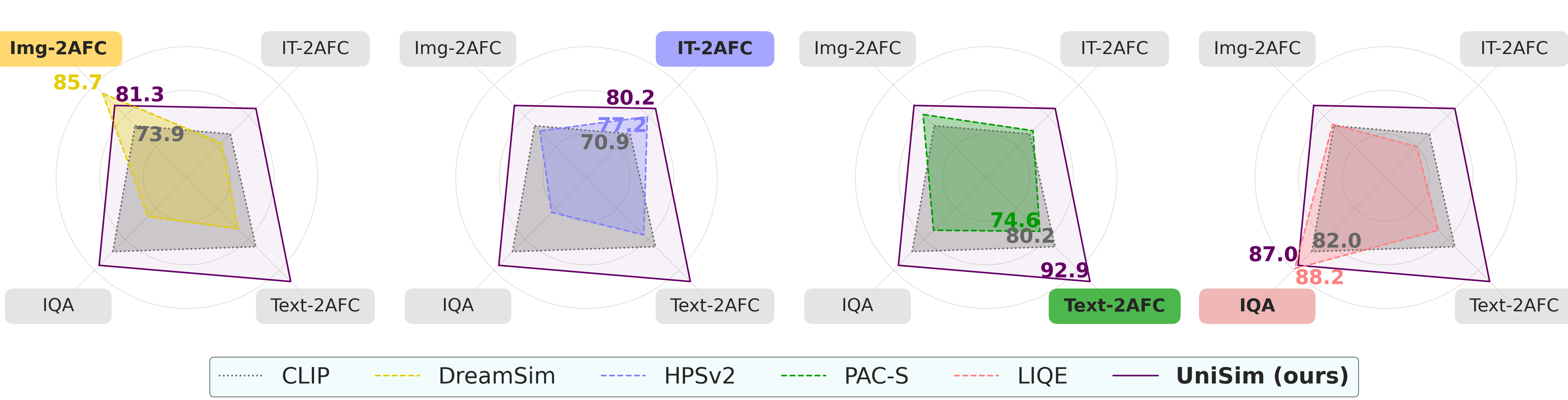}
  \captionof{figure}{\textbf{Summary of our UniSim framework.} First, we frame the existing multi-modal perceptual similarity tasks into our unified benchmark \textbf{\unisimbnc} (from which the \coretasks are illustrated in the top row).
  Second, we show that models specialized in individual tasks (e.g. DreamSim \cite{fu2023learning}, HPSv2 \cite{wu2023human}, PAC-S \cite{sarto2023positive}, LIQE \cite{zhang2023liqe}) do not generalize well to unseen perceptual tasks, with even worse accuracy than CLIP \cite{radford2021clip}.
  Finally, we introduce our multi-task perceptual metric, \textbf{UniSim}, which surpasses the baseline CLIP model and demonstrates superior or competitive performance compared with the specialized models, as depicted in the radar plots.
  }
  \label{fig:teaser}
\end{center}
}]
\blfootnote{$^{*}$Correspondence: \texttt{sg7457@nyu.edu}}
\begin{abstract}

Human perception of similarity across uni- and multi-modal inputs is highly complex, making it challenging to develop automated metrics that accurately mimic it.
General-purpose vision-language models, such as CLIP and large multi-modal models (LMMs), can be applied as zero-shot perceptual metrics, and several recent works have developed models specialized in narrow perceptual tasks.
However, the extent to which existing perceptual metrics align with human perception remains unclear.
To investigate this question, we introduce \unisimbnc, a benchmark encompassing 7 multi-modal perceptual similarity tasks, with a total of 25 datasets.
Our evaluation reveals that while general-purpose models perform reasonably well on average, they often lag behind specialized models on individual tasks.
Conversely,  metrics fine-tuned for specific tasks fail to generalize well to unseen, though related, tasks.
As a first step towards a unified multi-task perceptual similarity metric, we fine-tune both encoder-based and generative vision-language models on a subset of the \unisimbnc tasks.
This approach yields the highest average performance, and in some cases, even surpasses task-specific models.
Nevertheless, these models still struggle with generalization to unseen tasks, highlighting the ongoing challenge of learning a robust, unified perceptual similarity metric capable of capturing the human notion of similarity. The code and models are available at \href{https://github.com/SaraGhazanfari/UniSim}{github.com/SaraGhazanfari/UniSim}.
\end{abstract}

\vspace{-0.5cm}
\section{Introduction}
Developing automated metrics that replicate human perception of similarity remains a complex and open problem due to its intricate and multi-dimensional nature.

With the rapid progress of vision-language models \cite{radford2021clip,  podell2023sdxl, liu2024improved, li2024llava, achiam2023gpt}, there is a growing need for metrics that can evaluate similarity across multiple input modalities.
Effective perceptual metrics can benefit several perceptual applications, such as measuring the alignment of large autoregressive multi-modal models (LMMs) \cite{achiam2023gpt, team2024chameleon}, or evaluating the quality of text-to-image generative models \cite{zhang2023text}.
Moreover, human-aligned visual representations (achieved by fine-tuning the foundation models on perceptual data) have recently been shown to outperform non-aligned ones in certain vision (non-perceptual) downstream tasks~\cite{sundaram2024does}.

Prior works \cite{hessel2021clipscore, fu2023learning} have shown that foundation encoder models like CLIP \cite{radford2021clip} or DINO \cite{caron2021emerging} can be used as expressive metrics, where the semantic similarity between visual or text inputs 
is approximated through the alignment of embedding vectors.
Moreover, LMMs \cite{li2024llava, jiang2024mantis, achiam2023gpt} can be prompted to solve perceptual tasks using natural language.
While these models exhibit strong zero-shot performance on some perceptual tasks, they often struggle with more fine-grained or complex tasks.
As a result, specializing encoder-based \cite{ liu2021fusedream,lee2021umic, xu2023imagereward, wu2023human, sarto2023positive, wada2024polos, zhang2023liqe} and generative models \cite{wu2023q, wu2024towards, zhu2024adaptive} for narrow applications, e.g. image-to-image similarity or text-image alignment, has become a promising research direction.

Despite this progress, it remains unclear how well current metrics capture the complexity of human perception of similarity.  To investigate this, we argue that the various perceptual tasks addressed separately by specialized models represent distinct facets of the human perception system. Therefore, a unified framework is essential to holistically evaluate and integrate these aspects, enabling the development of metrics that better align with 
human perception.
Thus, in the first step,  we introduce \unisimbnc, a benchmark integrating  7 widely-used uni- and multi-modal perceptual tasks, encompassing 25 datasets, in a single framework (examples in Fig.~\ref{fig:teaser}).
In particular, we split the seven tasks into two categories: the first includes two-alternative forced choice (2AFC) tasks \cite{Link2AFC_1975}, offering a set of complementary uni- and multi-modal similarity tasks, while the second group is designed to assess perceptual understanding in more diverse scenarios.
Our evaluation of existing models on \unisimbnc reveals significant limitations: models fine-tuned for specific tasks often struggle to generalize across multiple perceptual tasks (Fig.~\ref{fig:teaser}), and even within different datasets of the same task. This shortcoming highlights the current gaps in their alignment with human perception and limits their practical applicability.

Then, to address these limitations, we propose  UniSim, a family of unified multi-task perceptual models. We fine-tune both CLIP \cite{radford2021clip} and LLaVA-NeXT \cite{li2024llava} using tailored multi-task learning approaches across a diverse set of perceptual datasets. The UniSim models achieve higher average accuracy compared to baselines across multiple tasks and exhibit generalization to left-out datasets within each task, showing the viability of a unified perceptual metric. For out-of-distribution (OOD) perceptual tasks, UniSim demonstrates improvements on unseen tasks which have structures similar to the training ones. However, generalizing to more diverse uni- and multi-modal tasks remains a significant challenge. This observation underscores the need for further research on learning perceptual similarity metrics that align broadly with human judgment.
\\

\noindent\textbf{Contributions.} In summary, the main contributions of our work, also illustrated in Fig.~\ref{fig:teaser}, are:

\begin{itemize}[parsep=3pt, topsep=0pt]
\item \textbf{\unisimbnc}, the first  benchmark to track the progress of perceptual similarity metrics across uni- and multi-modal tasks, including both core 2AFC tasks and a diverse set of tasks to test out-of-distribution generalization,
\item Identification of the limitations of current specialized perceptual metrics in generalizing to unseen datasets and perceptual tasks,  providing deeper insights into the shortcomings of existing  metrics,
\item \textbf{UniSim}, a set of multi-task perceptual models 
which are a first step towards general-purpose perceptual metrics.
\end{itemize} 

\noindent Together, \unisimbnc and UniSim lay the groundwork for understanding the challenges of learning automated metrics that broadly  mimic human perceptual similarity, beyond narrow, task-specific applications.

\section{Background and Related Work}
\label{sec:related_work}
In the following, we provide background on perceptual similarity tasks and metrics, along with the most relevant related works (an extended discussion is available in  App.~\ref{sec:extended_related_work}).

\subsection{Perceptual similarity tasks}

Learning to assess the similarity between data items in a way that aligns with human perception has long been a core challenge in computer vision and machine learning.
Traditional perceptual metrics often focused on uni-modal tasks, e.g. assessing image-to-image similarity \cite{zhang2018unreasonable, fu2023learning} or quality in denoising and compression contexts~\cite{wang2023exploring, zhang2023liqe}.
Recent advances in generative and multi-modal AI call however for perceptual metrics addressing cross-modal consistency, as they are used for training and evaluating text-to-image generative models \cite{xu2023imagereward, wu2023human}, captioning models \cite{li2023blip2, li2024llava}, and the perceptual capabilities of multi-modal LLMs \cite{li2024llava, jiang2024mantis}.
Despite shared goals, prior work has generally treated these perceptual tasks as isolated problems, and developed distinct approaches.
As a cohesive framework that integrates these tasks is still lacking, we propose a unified framework that 1) enables consistent evaluations of existing metrics, and 2) fosters the development of generalized perceptual similarity metrics across uni- and multi-modal domains.

\subsection{Foundation models as perceptual metrics}\label{sec:background_models}

\noindent\textbf{Encoder models.}
Replacing raw data with deep features extracted from pre-trained neural networks has become the standard in perceptual metrics, emphasizing learned representations over low-level comparisons. 
This approach leverages the high-level feature spaces within deep networks to better capture human-perceived similarity compared to traditional metrics, and is used in tasks like image-to-image similarity~\cite{zhang2018unreasonable, liu2021fusedream, croce2024adversarially}, text-image alignment \cite{hessel2021clipscore, lee2021umic, xu2023imagereward, sarto2023positive, wada2024polos}, image quality assessment \cite{wu2023human, zhang2023liqe}.
Foundation models like CLIP~\cite{radford2021clip} and BLIP~\cite{li2023blip2} have been the basis for many of these metrics.
Specifically, CLIP consists of an image encoder, $\phi: I \rightarrow \R^D$, and a text encoder $\psi: T \rightarrow \R^D$, which project data from different modalities into a shared $D$-dimensional latent space.
Using contrastive learning, CLIP aligns the embeddings of image-text pairs with their corresponding semantic meanings within this latent space. The similarity between inputs can be then quantified by the cosine similarity of their embedding vectors.
For instance, given a caption $\vt \in T$ and two images  $\vx_1, \vx_2 \in I$, a CLIP model can determine which image better aligns with the caption by solving: $\argmax_{\vz\in\{\vx_1, \vx_2\}} \simfn_{\phi, \psi}(\vz, \vt)$, where 
    $
    \simfn_{\phi, \psi}(\vz, \vt) = \inner{\frac{\phi(\vz)}{\norm{\phi(\vz)}_2},\frac{\psi(\vt)}{\norm{\psi(\vt)}_2}}
    \label{eq:clip_similarity}
    $
is the generic similarity function that uses the CLIP encoders $\phi, \psi$ to measure the similarity of the items of any input pair (in this case an image-text pair).
Encoder models have the advantage of associating each input with a single feature vector, allowing reuse for multiple comparisons. This leads to efficient evaluation across several images and texts without redundant computations. This efficiency is especially valuable in tasks like retrieval, where the similarity of a query must be measured against hundreds or thousands of references.
\\

\noindent\textbf{Generative models.}
Recently, large multi-modal models (LMMs) have made significant progress~\cite{liu2024improved,ye2024mplug,ghazanfari2024emma}, 
achieving strong capabilities in multi-image understanding and reasoning~\cite{li2024llava,jiang2024mantis,ye2024mplug}.  
This makes LMMs promising alternatives to traditional encoder models for perceptual metrics.
A generalist LMM can be easily adapted for specific perceptual tasks using simple prompting.
For instance, in the example above, one could query \texttt{``Image A: <$\vx_1$>, Image B: <$\vx_2$>. Which image is better described by <$\vt$>?''}.
This approach offers greater flexibility than encoder models, leveraging the extensive training and scale of these large models. However, a key drawback is the challenge of scaling LMMs to tasks involving many text prompts or images, such as image-to-image retrieval.
In addition to generalist models \cite{li2024llava, jiang2024mantis}, several works have specialized LMMs for specific perceptual tasks, often focusing on single-image evaluations, such as image quality assessment \cite{wu2023q, wu2024towards, zhu2024adaptive}, and  image aesthetics evaluation~\cite{huang2024aesexpert}.

\subsection{Benchmarks}
In recent years, several benchmarks have been developed to evaluate the perceptual and multi-modal understanding capabilities of large vision-language models. 
BLINK~\cite{fu2024blink} 
covers 14 visual perception tasks, but includes only a single dataset for image-to-image similarity. 
MUIRBENCH~\cite{wang2024muirbench} provides images and multiple-choice questions to assess 12 multi-image understanding tasks, including one that evaluates the image-text alignment. 
Also about image-text similarity, several benchmarks \cite{ku2023imagenhub, li2024aigiqa, li2023agiqa, zhang2024bench} offer comprehensive frameworks for evaluating text-to-image generative models. 
In the area of visual quality analysis, Q-Bench~\cite{wu2023q}, its enhanced version Q-Bench+,  2AFC-LMM~\cite{zhu20242afc}, and MICBench~\cite{wu2024towards}  assess a wide range of visual attributes, including low-level perception, detailed description, and overall quality.
While each of these benchmarks addresses some particular facets of perceptual evaluation, they often 
focus on reasoning and understanding tasks. 
This underscores further the need for a comprehensive benchmark to assess the quality of existing and new automated metrics across all aspects of multi-modal similarity perception.

\section{Towards a Unified Framework for Multi-Modal Perceptual Similarity Tasks}
\label{sec:our_benchmark}
We here introduce our unified framework for benchmarking and designing versatile perceptual similarity metrics. With \unisimbnc, we combine several perceptual tasks (Sec.~\ref{sec:tasks}) into a cohesive benchmark (Sec.~\ref{sec:unisim-bench}):
while diverse, we consider these tasks as specific instances of a broader challenge, i.e. capturing the human perception of similarity. 
Finally, in Sec.~\ref{sec:unisim-models} we leverage \unisimbnc to develop multi-task perceptual models, collectively named UniSim.

\subsection{Multi-modal perceptual similarity tasks} \label{sec:tasks}

\begin{figure}[t]
    \centering
    \includegraphics[width=\linewidth, trim=6mm 20mm 182mm 15mm, clip]{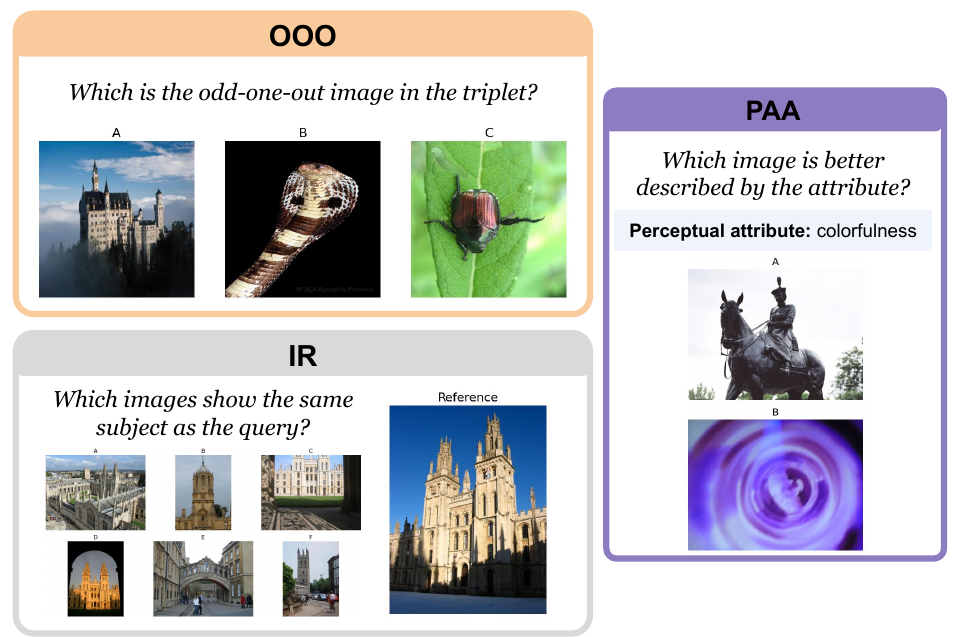}
    \vspace{-5mm}
    \caption{\textbf{\gentasks in \unisimbnc.} We illustrate samples from the three tasks not used for training, but to evaluate the model's generalization capabilities.}
    \label{fig:unseen_tasks}
    \vspace{-0.2cm}
\end{figure}

The following tasks, visualized in Fig.~\ref{fig:teaser} and Fig.~\ref{fig:unseen_tasks}, are the basis of our benchmark (further details in App.~\ref{sec:exp_details}).

\begin{description}[leftmargin=0pt,itemsep=3pt,topsep=3pt,parsep=3pt]
\item[Image-to-Image Similarity (\imgtwoafc).]
In this task, each data point consists of a triplet $(\vx_\textrm{ref}, \vx_1, \vx_2)$, and one has to decide which of two images $\vx_1, \vx_2$ is more similar to the reference image $\vx_\textrm{ref}$.
The \bapps \cite{zhang2018unreasonable} (used to tune the LPIPS metric) and \pieapp \cite{prashnani2018pieapp} datasets contain images perturbed with different corruptions, and compare their similarity to the original images. 
Conversely, \nights \cite{fu2023learning} includes high-resolution synthetic images, and aims at capturing more high-level similarity (pose, perspective, number of items, etc.).
The labels describe the human preference between the alternative images.

\item[Image-to-Text Alignment (\ittwoafc).]
Perceptual metrics are utilized to assess the 
synthetic images generated by text-to-image models, in terms of quality and alignment with the prompt. 
Here, the inputs are triplets $(\vt_\textrm{ref}, \vx_1, \vx_2)$, and the goal to decide which image $\vx_1, \vx_2$ better captures the prompt $\vt_\textrm{ref}$.
We rely on the \imgreward~\cite{xu2023imagereward}, \hpd~\cite{wu2023human}, \agiqa~\cite{li2023agiqa} datasets: for each text prompt, they comprise a set of synthetic images ranked (sorted or scored) by experts to reflect human preferences. 
Also, datasets for instruction-guided image editing, which include annotated source and target images along with textual instructions, align naturally with the \nittwoafc task. Thus, we include \magicbrush~\cite{zhang2024magicbrush} and \hqedit~\cite{hui2024hq}.

\item[Text-to-Image Alignment (\texttwoafc).]
Assessing the quality and specificity of generated captions for a given image is essential for ensuring accurate and meaningful text generation.
Thus, we incorporate the \ntexttwoafc task, which can be seen as the reverse of \nittwoafc, where the goal is to select the text $\vt_1$ or  $\vt_2$ that better describes the reference image $\vx_\textrm{ref}$.
We use three datasets for this task: \polaris~\cite{wada2024polos}, \coco~\cite{bianco2023improving} (based on the \mscoco~\cite{lin2014microsoft} images) and \hqedit~\cite{hui2024hq}.
\item[Image Quality Assessment (\iqa).]
In this well-established task, one has to determine which of two images $\vx_1, \vx_2$ has higher quality.
While some works focus on no-reference quality assessment (i.e., assigning an absolute score), we limit our evaluation to pairwise comparisons.
The \kadid dataset \cite{hanhe2019kadid10k} contains images artificially corrupted with varying levels of severity, while \koniq \cite{hosu2020koniq10k} includes images with authentic distortion.
Moreover, we again use \pieapp \cite{prashnani2018pieapp} 
and \agiqa~\cite{li2023agiqa} which also provide quality scores.
Finally, we include \pipal~\cite{jinjin2020pipal} which contains high-quality reference images subjected to 116 types of distortions.

\item[Perceptual Attributes Assessment (\paa).]
Here, we evaluate specific visual characteristics (attributes) of the image that directly affect human perception and contribute to overall visual quality.
The perceptual attributes considered are \textit{brightness}, \textit{colorfulness}, \textit{contrast}, and \textit{sharpness}.
For an image pair $(\vx_1, \vx_2)$, the task is to choose which one better exhibits the given attribute (e.g., which is brighter).
We use the \koniq~\cite{hosu2020koniq10k} dataset for all attributes and additionally leverage the \sice~\cite{cai2018learning} dataset for brightness.

\item[Odd-One-Out (\ooo).]
Given a triplet of images $(\vx_1, \vx_2, \vx_3)$, the task consists of finding the one that does not belong with the others---that is, the most dissimilar image.
We use the dataset derived by \cite{muttenthaler2023human} from the coarse \cifarh classes, named \cifarhooo.
Moreover, we follow a similar approach to obtain \imgnet.
We get 6 macro-classes by merging certain \imagenet-1k classes, which yields semantically distinct classes with sufficient intra-class diversity, making the task non-trivial.
A triplet consists of two images from the same macro-class and one from a different one, the ground-truth odd one.

\item[Image-to-Image Retrieval (\ir).]
Perceptual metrics have also been employed to identify the closest matches to a query image within a database.
Unlike the previous tasks, retrieval involves ranking the entire pool of images based on similarity to the query rather than choosing between 2-3 alternatives.
Nevertheless, we consider it a relevant use case for perceptual similarity metrics and include it in our framework.
We employ the  \oxford and \paris datasets~\cite{oxfordparis} 
(medium and hard difficulty levels).

\end{description}

\subsection{\unisimbnc: an open-ended multi-modal perceptual similarity benchmark} \label{sec:unisim-bench}

Building on the 
multi-modal perceptual tasks from Sec.~\ref{sec:tasks}, we now 
present our unified framework \unisimbnc.
\\

\noindent\textbf{Composition.}
We split the tasks from Sec.~\ref{sec:tasks} into two groups: 
the first group consists of the \textbf{\coretasks}—\nimgtwoafc, \nittwoafc, \ntexttwoafc, and \niqa---which form a diverse set of 
complementary tasks to evaluate different aspects of perceptual similarity.
These tasks are well-established with numerous datasets available that can be framed as 2AFC problems. Each task also has multiple specialized models, providing strong baselines for our evaluation.
The second group consists of the  \textbf{\gentasks}, including \npaa, \nooo, and \nir, which capture more peripheral yet important aspects of perception. These tasks assess how well perceptual metrics trained on  core tasks generalize to new unseen perceptual challenges, and can be seen as out-of-distribution (OOD) compared to the core tasks.
Together, the two splits form \unisimbnc, which includes 7 tasks and 25 datasets (we provide in App.~\ref{sec:exp_details} more details on the resulting benchmark).
\\

\noindent\textbf{Discussion.}
In designing our benchmark, we aimed to capture a wide range of perceptual similarity tasks to enable a comprehensive evaluation of existing automated metrics.
While this set, to the best of our knowledge, forms the broadest benchmark currently available for the topic, we consider it an open-ended effort.
Future expansions could include additional applications of perceptual similarity metrics and higher-quality datasets for existing tasks.
Despite potential limitations, we believe our benchmark provides valuable insights into the shortcomings of current metrics and offers a foundation for the development of more robust metrics across diverse modalities and applications, as explored in the following sections.

\subsection{UniSim: A Family of Multi-task Perceptual Similarity Metrics} \label{sec:unisim-models}

We now introduce UniSim, our unified multi-task perceptual metric
which integrates
several perceptual tasks in a multi-modal metric. UniSim is trained on a subset of  datasets from the core tasks of \unisimbnc, as detailed in 
App.~\ref{sec:exp_details}.
In particular, the tasks from the \gentasks category are entirely excluded from training. 
Additionally, certain datasets from the \textit{\coretasks} (i.e., \bapps, \imgreward, \agiqa, \coco, \koniq) are deliberately withheld for evaluating generalization. 
This approach allows us to assess three types of generalization of perceptual metrics: 1) standard \textit{training-test} set generalization (for datasets included in training), 2) \textit{intra-task} generalization, where the model is tested on unseen datasets from known (core) tasks, 3) \textit{inter-task} generalization, where the model is tested on new, unseen (non-2AFC) tasks. 
Structuring the training data as two-alternative forced choice tasks offers two practical advantages. 
First, it leverages the efficiency of encoder models like CLIP, requiring only pairwise comparisons and allowing the problem to be framed as a binary classification task. Second, these tasks involve at most three images, simplifying the training process for large multi-modal models. 
With this setup, we fine-tune UniSim models from either CLIP models or LMMs, so to leverage the extensive knowledge accumulated during their pre-training on large-scale multi-modal datasets, as we describe below (additional information about models and training schemes in App.~\ref{sec:exp_details}).
\\

\noindent\textbf{CLIP-based UniSim.}
For training on multiple tasks, a significant challenge is unifying datasets with different formats.
Thanks to the standardized 2AFC structure of the core tasks in \unisimbnc, each data point is a triplet $(\vz_\textrm{ref}, \vz_0, \vz_1)$ consisting of text prompts or images, with a reference item $\vz_\textrm{ref}$ and two alternatives  $\vz_0, \vz_1$, as well as a label $y\in\{0, 1\}$ indicating which alternative is more similar to the reference.\footnote{For \niqa we use the prompt \texttt{``A high quality photo.''} as reference to complete the triplet} 
This setup allows us to frame the problem as  binary classification, and fine-tune a CLIP model to solve it by optimizing,  as in earlier methods~\cite{liu2021fusedream}, the hinge loss
\begin{align}
\begin{split}
    \L(&\vz_\textrm{ref}, \vz_0, \vz_1, y, \phi, \psi) =  \max\{0, (2y - 1) \cdot \\ &(\simfn_{\phi, \psi}(\vz_\textrm{ref}, \vz_0) - \simfn_{\phi, \psi}(\vz_\textrm{ref}, \vz_1)) + \mu\},
    \end{split}
    \label{eq:hinge_loss}
\end{align}
where $\simfn_{\phi, \psi}$ is the similarity function induced by the CLIP model (with encoders $\phi, \psi$, see 
Sec.~\ref{sec:background_models}), and $\mu \geq 0$ a margin to ensure confident predictions.
We fine-tune only the image encoder $\phi$, i.e. the text encoder $\psi$ is frozen. 
We concatenate the datasets belonging to the same task, and denote the $i$-th data sample for the $t$-th tasks  as  $(\vz_\textrm{ref}^{(t, i)}, \vz_0^{(t, i)}, \vz_1^{(t, i)}, y^{(t, i)})$:
this yields the training objective
\vspace{-3mm}
\begin{align}
    \min_\phi \; \sum_{t=1}^{4} \, \sum_{i=1}^n \L(\vz_\textrm{ref}^{(t, i)}, \vz_0^{(t, i)}, \vz_1^{(t, i)}, y^{(t, i)}, \phi, \psi)
    \label{eq:multi-task-loss}
\end{align}
where, in practice, we replace $n$ (the entire dataset) with the batch size used for training.
This approach ensures that the number of samples seen is balanced across tasks, regardless of the dataset size. 
We use LoRA
~\cite{hu2022lora}, following~\cite{liu2021fusedream,croce2024adversarially}, to enable efficient fine-tuning while mitigating overfitting.
We apply this approach to the CLIP model with ViT-B/32 as vision encoder from the OpenCLIP library \cite{cherti2023openclip} and to the original CLIP (ViT-L/14 backbone, 336x336 input resolution) from \cite{radford2021clip}, resulting in two encoder-based UniSim models of different size.
\\
\begin{table*}[h]
\centering \small
\tabcolsep=1.2pt
\extrarowheight=1.5pt
\myindent=4mm
\newl=7mm
\newcommand{\myangle}{90}
\definecolor{myTableGray}{rgb}{.56, .56, .56}

\scalebox{.885}{
\begin{tabular}{L{29mm}||
*{3}{C{\newl}}
>{\columncolor{myYellow!50}}C{\newl}|
*{5}{C{\newl}}
>{\columncolor{myBlue!50}}C{\newl}|
*{3}{C{\newl}}
>{\columncolor{myGreen!60}}C{\newl}|
*{5}{C{\newl}}
>{\columncolor{myRed!50}}C{\newl}||
>{\columncolor{myGray!50}}C{\newl}
}
\toprule

& \multicolumn{4}{c}{\cellcolor{myYellow} \textbf{\imgtwoafc}}  & \multicolumn{6}{c}{\cellcolor{myBlue} \textbf{\ittwoafc}}  & \multicolumn{4}{c}{\cellcolor{myGreen} \textbf{\texttwoafc}}  & \multicolumn{6}{c}{\cellcolor{myRed} \textbf{\iqa}}  & \textbf{Avg} \\

Models  & 
\rotatebox{\myangle}{\nights$^{(1,\dagger)}$} &
\rotatebox{\myangle}{\bapps} &
\rotatebox{\myangle}{\pieapp$^{(\dagger)}$}  & 
\rotatebox{\myangle}{\textbf{average}} &
\rotatebox{\myangle}{\imgreward$^{(3)}$} &
\rotatebox{\myangle}{\hpd$^{(4, \dagger)}$} &
\rotatebox{\myangle}{\agiqa} &
\rotatebox{\myangle}{\magicbrush$^{(\dagger)}$} &
\rotatebox{\myangle}{\hqedit$^{(\dagger)}$}  & 
\rotatebox{\myangle}{\textbf{average}} &
\rotatebox{\myangle}{\coco} &
\rotatebox{\myangle}{\polaris$^{(\dagger)}$} &
\rotatebox{\myangle}{\hqedit$^{(\dagger)}$}  & 
\rotatebox{\myangle}{\textbf{average}} &
\rotatebox{\myangle}{\kadid$^{(5, \dagger)}$} &
\rotatebox{\myangle}{\koniq$^{(6)}$} &
\rotatebox{\myangle}{\pieapp$^{(\dagger)}$} &
\rotatebox{\myangle}{\agiqa} &
\rotatebox{\myangle}{\pipal$^{(\dagger)}$}  & 
\rotatebox{\myangle}{\textbf{average}} &
\\
\toprule
\multicolumn{10}{l}{\textbf{General-purpose models}}\\
\toprule

\textbf{CLIP} ViT-B/32 & \textcolor{myTableGray}{85.1}  & \textcolor{myTableGray}{68.6}  & \textcolor{myTableGray}{80.2}  & 78.0  & \textcolor{myTableGray}{65.8}  & \textcolor{myTableGray}{63.3}  & \textcolor{myTableGray}{66.1}  & \textcolor{myTableGray}{72.4}  & \textcolor{myTableGray}{85.2}  & 70.6 & \textcolor{myTableGray}{61.4}  & \textcolor{myTableGray}{78.9}  & \textcolor{myTableGray}{84.6}  & 75.0 &
\textcolor{myTableGray}{59.8} & \textcolor{myTableGray}{51.8} & \textcolor{myTableGray}{80.5} & \textcolor{myTableGray}{68.3} & \textcolor{myTableGray}{74.4} &	67.0 & 72.6\\

\textbf{CLIP} ViT-L/14 & \textcolor{myTableGray}{81.5} & \textcolor{myTableGray}{64.2} & \textcolor{myTableGray}{76.1} & 73.9  & \textcolor{myTableGray}{63.1} & \textcolor{myTableGray}{65.8} & \textcolor{myTableGray}{62.9} & \textcolor{myTableGray}{78.2} & \textcolor{myTableGray}{84.7} & 70.9  & \textcolor{myTableGray}{75.0} & \textcolor{myTableGray}{82.0} & \textcolor{myTableGray}{83.6} & 80.2  & \textcolor{myTableGray}{84.1} & \textcolor{myTableGray}{69.1} & \textcolor{myTableGray}{90.5} & \textcolor{myTableGray}{77.7} & \textcolor{myTableGray}{88.8} & 82.0  & 76.8 \\

\textbf{CLIP} ViT-H/14  & \textcolor{myTableGray}{84.0} & \textcolor{myTableGray}{69.0} & \textcolor{myTableGray}{76.8} & 76.6  & \textcolor{myTableGray}{63.3} & \textcolor{myTableGray}{65.5} & \textcolor{myTableGray}{65.1} & \textcolor{myTableGray}{76.5} & \textcolor{myTableGray}{\underline{86.5}} & 71.4  & \textcolor{myTableGray}{66.4} & \textcolor{myTableGray}{81.8} & \textcolor{myTableGray}{85.6} & 77.9  & \textcolor{myTableGray}{67.0} & \textcolor{myTableGray}{61.1} & \textcolor{myTableGray}{72.0} & \textcolor{myTableGray}{65.7} & \textcolor{myTableGray}{67.5} &	66.7  & 73.1\\

\textbf{BLIP} ViT-L/14  & \textcolor{myTableGray}{80.8} & \textcolor{myTableGray}{65.0} & \textcolor{myTableGray}{72.1} & 72.6  & \textcolor{myTableGray}{64.1} & \textcolor{myTableGray}{67.0} & \textcolor{myTableGray}{64.5} & \textcolor{myTableGray}{73.3} & \textcolor{myTableGray}{85.4} & 70.9  & \textcolor{myTableGray}{66.3} & \textcolor{myTableGray}{78.9} & \textcolor{myTableGray}{82.6} & 75.9  & \textcolor{myTableGray}{65.1} & \textcolor{myTableGray}{55.2} & \textcolor{myTableGray}{61.0} & \textcolor{myTableGray}{57.0} & \textcolor{myTableGray}{61.9} & 60.0  & 69.9 \\


\textbf{LLaVA-NeXT}-0.5B$^{\clubsuit}$  & \textcolor{myTableGray}{58.7} & \textcolor{myTableGray}{52.8} & \textcolor{myTableGray}{63.0} & 58.2  & \textcolor{myTableGray}{61.3} & \textcolor{myTableGray}{76.6} & \textcolor{myTableGray}{65.2} & \textcolor{myTableGray}{64.4} & \textcolor{myTableGray}{75.1} & 68.5  & \textcolor{myTableGray}{53.7} & \textcolor{myTableGray}{71.6} & \textcolor{myTableGray}{57.9} & 61.1  & \textcolor{myTableGray}{53.6} & \textcolor{myTableGray}{52.7} & \textcolor{myTableGray}{55.1} & \textcolor{myTableGray}{57.5} & \textcolor{myTableGray}{50.8} & 53.9  & 60.4 \\

\textbf{LLaVA-NeXT}-7B$^{\clubsuit}$  & \textcolor{myTableGray}{\underline{91.3}} & \textcolor{myTableGray}{67.0} & \textcolor{myTableGray}{79.9} & 79.4  & \textcolor{myTableGray}{71.5} & \textcolor{myTableGray}{76.1} & \textcolor{myTableGray}{\underline{68.5}} & \textcolor{myTableGray}{72.7} & \textcolor{myTableGray}{\underline{86.5}} & 75.1  & \textcolor{myTableGray}{59.6} & \textcolor{myTableGray}{79.4} & \textcolor{myTableGray}{80.0} & 73.0  & \textcolor{myTableGray}{66.1} & \textcolor{myTableGray}{79.2} & \textcolor{myTableGray}{83.6} & \textbf{\textcolor{myTableGray}{80.6}} & \textcolor{myTableGray}{80.9} & 78.1  & 76.4 \\

\textbf{Mantis} Idefics-8B$^{\clubsuit}$  & \textcolor{myTableGray}{89.5} & \textcolor{myTableGray}{63.8} & \textcolor{myTableGray}{75.0} & 76.1  & \textcolor{myTableGray}{71.0} & \textcolor{myTableGray}{73.9} & \textcolor{myTableGray}{68.5} & \textcolor{myTableGray}{75.8} & \textcolor{myTableGray}{84.4} & 74.7  & \textcolor{myTableGray}{64.7} & \textcolor{myTableGray}{77.8} & \textcolor{myTableGray}{83.0} & 75.2  & \textcolor{myTableGray}{58.3} & \textcolor{myTableGray}{76.3} & \textcolor{myTableGray}{65.1} & \textcolor{myTableGray}{79.0} & \textcolor{myTableGray}{74.9} & 70.7  & 74.2 \\
\midrule

\multicolumn{21}{l}{\textbf{Specialized models}} \\
\toprule
\cellcolor{myYellow!50}\textbf{DS}$^{(1)}$ ViT-B/32  & \textbf{\textcolor{myTableGray}{95.3}} & \textbf{\textcolor{myTableGray}{73.3}} & \textbf{\textcolor{myTableGray}{88.5}} & \textbf{85.7} & \textcolor{myTableGray}{63.1} & \textcolor{myTableGray}{62.0} & \textcolor{myTableGray}{64.4} & \textcolor{myTableGray}{68.8} & \textcolor{myTableGray}{79.8} & 67.6  & \textcolor{myTableGray}{61.3} & \textcolor{myTableGray}{75.6} & \textcolor{myTableGray}{84.1} & 73.7  & \textcolor{myTableGray}{70.1} & \textcolor{myTableGray}{58.0} & \textcolor{myTableGray}{78.4} & \textcolor{myTableGray}{67.1} & \textcolor{myTableGray}{72.7} & 69.2  & 74.1 \\


\cellcolor{myBlue!60}\textbf{IR}$^{(3)}$ BLIP 
& \textcolor{myTableGray}{87.1} & \textcolor{myTableGray}{66.1} & \textcolor{myTableGray}{77.6} & 76.9 & \textbf{\textcolor{myTableGray}{74.3}} & \textcolor{myTableGray}{74.5} & \textcolor{myTableGray}{\underline{72.4}} & \textcolor{myTableGray}{74.3} & \textcolor{myTableGray}{83.5} & 75.8  & \textcolor{myTableGray}{54.2} & \textcolor{myTableGray}{72.2} & \textcolor{myTableGray}{85.4} & 70.6  &

\textcolor{myTableGray}{62.3} & \textcolor{myTableGray}{58.0} & \textcolor{myTableGray}{75.1} & \textcolor{myTableGray}{74.8} & \textcolor{myTableGray}{60.1} & 66.1 & 72.3
\\

\cellcolor{myBlue!60}\textbf{HPSv2}$^{(4)}$ ViT-H/14  & \textcolor{myTableGray}{78.5} & \textcolor{myTableGray}{66.7} & \textcolor{myTableGray}{70.8} & 72.0  & \textcolor{myTableGray}{\underline{73.8}} & \textbf{\textcolor{myTableGray}{83.5}} & \textbf{\textcolor{myTableGray}{72.6}} & \textcolor{myTableGray}{74.9} & \textcolor{myTableGray}{81.2} & 77.2  & \textcolor{myTableGray}{68.2} & \textcolor{myTableGray}{78.1} & \textcolor{myTableGray}{81.5} & 75.9  & 
\textcolor{myTableGray}{67.0} & \textcolor{myTableGray}{63.6} & \textcolor{myTableGray}{68.9} & \textcolor{myTableGray}{65.4} & \textcolor{myTableGray}{73.5} & 67.7  & 73.2\\

\cellcolor{myGreen!50}\textbf{PAC-S} ViT-L/14  & \textcolor{myTableGray}{86.9} & \textcolor{myTableGray}{69.1} & \textcolor{myTableGray}{78.1} & 78.0 & \textcolor{myTableGray}{65.0} & \textcolor{myTableGray}{67.0} & \textcolor{myTableGray}{65.8} & \textcolor{myTableGray}{75.6} & \textcolor{myTableGray}{86.9} & 72.1  & \textcolor{myTableGray}{60.5} & \textcolor{myTableGray}{77.6} & \textcolor{myTableGray}{85.6} & 74.6  & \textcolor{myTableGray}{75.0} & \textcolor{myTableGray}{56.5} & \textcolor{myTableGray}{86.1} & \textcolor{myTableGray}{70.0} & \textcolor{myTableGray}{83.2} & 74.2  & 74.7 \\

\cellcolor{myRed!50}\textbf{LIQE$^{(5,6)}$} ViT-B/32   
& \textcolor{myTableGray}{77.9} & \textcolor{myTableGray}{68.7} & \textcolor{myTableGray}{76.6} & 74.4 & \textcolor{myTableGray}{61.9} & \textcolor{myTableGray}{67.3} & \textcolor{myTableGray}{64.1} & \textcolor{myTableGray}{59.9} & \textcolor{myTableGray}{78.3} & 66.3 & \textcolor{myTableGray}{63.5} & \textcolor{myTableGray}{78.2} & \textcolor{myTableGray}{81.0} & 74.2 & 
\textcolor{myTableGray}{92.4} & \textcolor{myTableGray}{\underline{87.9}} & \textcolor{myTableGray}{98.2} & \textcolor{myTableGray}{76.7} & \textcolor{myTableGray}{86.0} & \underline{88.2} & 75.8
\\

\cellcolor{myRed!50}\textbf{C2S$^{\clubsuit}$$^{(5,6)}$} mOwl-2 & \textcolor{myTableGray}{-} & \textcolor{myTableGray}{-} & \textcolor{myTableGray}{-} & -  & \textcolor{myTableGray}{-} & \textcolor{myTableGray}{-} & \textcolor{myTableGray}{-} & \textcolor{myTableGray}{-} & \textcolor{myTableGray}{-} & -  & \textcolor{myTableGray}{-} & \textcolor{myTableGray}{-} & \textcolor{myTableGray}{-} & -  & \textbf{\textcolor{myTableGray}{96.2}} & \textbf{\textcolor{myTableGray}{92.0}} & \textbf{\textcolor{myTableGray}{99.2}} & \textcolor{myTableGray}{76.3} & \textcolor{myTableGray}{87.3} & \textbf{90.2}  & - \\

\midrule

\multicolumn{21}{l}{\textbf{Our models}$^{(\dagger)}$} \\
\toprule
\textbf{UniSim} ViT-B/32  & \textcolor{myTableGray}{87.7} & \textcolor{myTableGray}{69.9} & \textcolor{myTableGray}{84.6} & 80.7  & \textcolor{myTableGray}{70.4} & \textcolor{myTableGray}{74.5} & \textcolor{myTableGray}{71.7} & \textcolor{myTableGray}{78.1} & \textcolor{myTableGray}{84.1} & 75.8  & \textcolor{myTableGray}{\underline{91.2}} & \textcolor{myTableGray}{94.2} & \textcolor{myTableGray}{85.6} & \underline{90.3}  & \textcolor{myTableGray}{89.9} & \textcolor{myTableGray}{72.0} & \textcolor{myTableGray}{93.6} & \textcolor{myTableGray}{77.3} &	\textcolor{myTableGray}{\textbf{93.4}} & 85.3  & 83.0 \\

\textbf{UniSim} ViT-L/14  & \textcolor{myTableGray}{90.7} & \textcolor{myTableGray}{68.1} & \textcolor{myTableGray}{85.0} & 81.3  & \textcolor{myTableGray}{69.4} & \textcolor{myTableGray}{\underline{82.3}} & \textcolor{myTableGray}{71.3} & \textbf{\textcolor{myTableGray}{91.8}} & \textcolor{myTableGray}{86.0} & \textbf{80.2}  & \textbf{\textcolor{myTableGray}{94.2}} & \textcolor{myTableGray}{\underline{96.1}} & \textcolor{myTableGray}{\underline{88.3}} & \textbf{92.9}  & \textcolor{myTableGray}{\underline{94.7}} & \textcolor{myTableGray}{71.8} & \textcolor{myTableGray}{\underline{98.9}} & \textcolor{myTableGray}{\underline{80.2}} & \textcolor{myTableGray}{89.2} & 87.0  & \textbf{85.3} \\

\textbf{UniSim}$^{\clubsuit}$ LL-N-0.5B  & \textcolor{myTableGray}{89.8} & \textcolor{myTableGray}{\underline{70.0}} & \textcolor{myTableGray}{\underline{85.3}} & \underline{81.7}  & \textcolor{myTableGray}{69.2} & \textcolor{myTableGray}{80.7} & \textcolor{myTableGray}{66.7} & \textcolor{myTableGray}{\underline{90.8}} & \textbf{\textcolor{myTableGray}{92.7}} & \underline{80.0} & \textcolor{myTableGray}{75.4} & \textbf{\textcolor{myTableGray}{99.9}} & \textbf{\textcolor{myTableGray}{89.2}} & 88.2  & \textcolor{myTableGray}{94.3} & \textcolor{myTableGray}{77.6} & \textcolor{myTableGray}{97.0} & \textcolor{myTableGray}{\textbf{80.6}} & \underline{\textcolor{myTableGray}{89.8}} & 87.9  & \underline{84.4} \\



\bottomrule
\end{tabular}}
\caption{\textbf{Evaluation on the \textit{Core 2AFC Tasks} of \unisimbnc.} We provide a comparative analysis of general-purpose, specialized, and UniSim models on the first section of \unisimbnc. LMM-based models are distinguished with the ${\clubsuit}$ symbol, while models highlighted with color are specialized in individual tasks (e.g., DS is specialized for the \imgtwoafc task).
Additionally, the datasets used for training each model are indicated as superscripts next to their names. \textbf{Observations:} (1) Specialized models generally perform worse than general-purpose models on tasks outside their training domain, highlighting a significant lack of generalization. For example, the HPSv2 model, which is specialized for the \ittwoafc task, performs worse than the baseline (ViT-H/14) on the closely related \texttwoafc task. (2) UniSim ranks as the first or second best across nearly all tasks, demonstrating the feasibility of training a unified multi-modal metric capable of handling diverse and widely-used tasks.
}
\vspace{-4mm}
\label{tab:unisim-result}
\end{table*}

\noindent\textbf{LMM-based UniSim.}
For the LMM-based version of our perceptual metric, we fine-tune the LLaVA-NeXT-0.5B model \cite{li2024llava}, as it has shown advanced capability to handle multi-image inputs and image-text interleaved formats.
Moreover, LLaVA-NeXT-0.5B has a relatively small number of parameters, making it significantly more efficient for training and inference---an important factor for real-world deployment.  
For the training, we leverage the instruction fine-tuning mechanism of LLaVA-NeXT-0.5B, and format our datasets to produce the annotation files compatible with those in \cite{li2024llava}, where the tasks are described as natural language instructions (see Sec.~\ref{sec:background_models}).
A significant challenge in fine-tuning LMMs for 2AFC tasks is that the ground truth consists of a single word representing the model's prediction between two alternatives.
To mitigate the risk of overfitting to specific structural patterns, 1) we design a variety of templates for both instructions and answers,  
ensuring diversity in the training data, and 2) we combine the Multi-image (500K) part of M4-Instruct~\cite{li2024llava} dataset with our perceptual dataset (842K). Notably, the datasets we use can be seamlessly integrated with any instruction-tuning dataset to fine-tune LMMs, enhancing their fine-grained perceptual capabilities.

\section{Evaluation on \unisimbnc}

Next, we use \unisimbnc for a comprehensive analysis of general-purpose, 
specialized, and our UniSim perceptual models, 
across perceptual tasks.
We first introduce the baselines  
(Sec.~\ref{sec:baselines}), then discuss the results on \coretasks (Sec.~\ref{sec:exp_core_tasks}) and \gentasks (Sec.~\ref{sec:exp_unseen_tasks}).

\subsection{Baselines} \label{sec:baselines}

\textbf{General-purpose multi-modal models.}
For encoder models, we benchmark the CLIP models with ViT-B/32, (which serves as the baseline for both DreamSim, LIQE, and UniSim-ViT-B/32), ViT-L/14 (baseline for PAC-S and UniSim-ViT-L/14), as well as ViT-H/14 (baseline for the HPS-v2 model).
We further test SigLIP SoViT-400m/14~\cite{alabdulmohsin2024getting} (results in appendix), and BLIP-2~\cite{li2023blip2} (with a ViT-L/14 encoder), which is the base model for ImageReward.
Among LMMs we include Llava-NeXT-0.5B~\cite{li2024llava} (basis of the LLM-based UniSim), its larger version Llava-NeXT-7B~\cite{li2024llava}, and the recent Mantis Idefics2-8B~\cite{jiang2024mantis}, a strong multi-image autoregressive model which can even match the performance of GPT-4V~\cite{achiam2023gpt} on multi-image tasks.\\

\noindent\textbf{Specialized perceptual metrics.}
For \nimgtwoafc, DreamSim (DS) \cite{fu2023learning} achieves SOTA performance via an ensemble of multiple vision encoders fine-tuned on \nights: since this is not associated with a text-encoder, we primarily compare their single-encoder (ViT-B/32) version.
For the \nittwoafc task, we select the ImageReward (IR) model \cite{xu2023imagereward} 
and HPSv2 \cite{wu2023human}: these are trained on the \imgreward and \hpd datasets respectively for evaluating text-to-image generative models.
As a metric specialized in \ntexttwoafc, we report the results of PAC-S \cite{sarto2023positive}, designed for image captioning evaluation.
Finally, for \niqa we report LIQE \cite{zhang2023liqe} and Compare2Score \cite{wu2024towards} (fine-tuned from mPLUG-Owl2-8B~\cite{ye2024mplug}) as encoder and generative baseline models respectively. We provide more details on the models in App.~\ref{sec:exp_details}, and the evaluation of additional baselines in App.~\ref{sec:additional_experiments}.

\subsection{Evaluation on \textbf{\coretasks}} \label{sec:exp_core_tasks}

\noindent\textbf{Evaluation setup.}
In Table~\ref{tab:unisim-result} we report the results of the various perceptual metrics on the core tasks of \unisimbnc in terms of average classification accuracy. 
Besides the performance of each dataset, we compute the performance per task (average of the accuracy on the individual datasets within the same task) in the column with background color, and the overall average (last column), i.e., the mean of the single-task average performance values.
Finally, when evaluating tasks beyond the specialized models’ original scope, we adapt them by directly utilizing their image and text encoders, as they cannot be applied in their default configurations.\\

\noindent \textbf{Intra-task generalization.}
Among the three tiers of generalization that \unisimbnc aims to evaluate, the standard \textit{training-test} set generalization is typically achieved by all specialized models and versions of UniSim. However, \textit{intra-task} generalization—where models are tested on unseen datasets within 
their training tasks—poses a significant challenge for most specialized models. While some models, such as DreamSim and LIQE, exhibit this capability, others fall short. For example, both HPSv2 and ImageReward perform worse than the generalist baselines on \hqedit, highlighting that existing approaches still struggle with intra-task generalization. Conversely, 
the UniSim models successfully generalize to the intra-tasks datasets and outperform the baseline on the left-out datasets, sometimes of a large margin e.g. on \coco.\\

\noindent\textbf{Inter-task generalization.}
First, Table~\ref{tab:unisim-result} indicates that models specialized for a single perceptual task often suffer performance degradation on tasks outside their training domain, as also illustrated in Fig.~\ref{fig:teaser}.
For instance, DreamSim achieves the best performance on \nimgtwoafc but underperforms compared to CLIP on \nittwoafc and \ntexttwoafc. This is likely due to overfitting to a single perceptual task. Additionally, fine-tuning on a vision-only task may adversely impact the alignment between image and text embeddings.
Similarly, HPSv2, specialized for \nittwoafc, underperforms compared to the baseline (CLIP with ViT-H/14) on \ntexttwoafc, highlighting a lack of generalization even across closely related tasks.
In contrast, UniSim consistently ranks as the first or second best across nearly all tasks and achieves the best average performance. It also improves upon the models from which it was fine-tuned across all tasks. Although this performance may be expected given its training on multiple datasets, it demonstrates the feasibility of developing a unified multi-modal metric that can effectively handle diverse, widely-used tasks.

\begin{table}[t]
\centering \small
\tabcolsep=1.2pt
\extrarowheight=1.5pt
\myindent=4mm
\newl=11mm
\newcommand{\myangle}{90}
\scalebox{.9}{
\begin{tabular}{L{37mm}|
*{3}{|C{\newl}}
||
>{\columncolor{myGray!50}}C{\newl}
}
\toprule

Models & \cellcolor{myPurple} \textbf{\paa} & \cellcolor{myOrange} \textbf{\ooo} & \cellcolor{myGray} \textbf{\ir} & \textbf{Avg} \\
\toprule

\multicolumn{5}{l}{\textbf{General-purpose models}}\\
\toprule
\textbf{CLIP} ViT-B/32  & 70.6 & \textbf{71.3} &  43.8 & \textbf{61.9}\\  
\textbf{CLIP} ViT-L/14 & 66.8 & 65.8  & 45.5 & 59.4\\ 
\textbf{CLIP} ViT-H/14 & 68.2 &  70.3& \textbf{50.2} & 62.9 \\ 
\textbf{BLIP} ViT-L/14 & 66.1 & 63.4 & 35.9 & 55.1\\
\textbf{LLaVA NeXT}-0.5B$^{\clubsuit}$ & 63.9 & 33.7 & - & -\\ 
\textbf{LLaVA NeXT}-7B$^{\clubsuit}$ & 68.3 & 61.0 & -& -\\ 
\textbf{Mantis} Idefics-8B$^{\clubsuit}$ & 68.2 & 44.1 & -& -\\ 
\toprule

\multicolumn{5}{l}{\textbf{Specialized models}}\\
\toprule
\cellcolor{myYellow!50}\textbf{DreamSim} 
ViT-B/32 & 70.7 & 61.4 & 38.0 & 56.6\\ 
\cellcolor{myBlue!60} \textbf{ImageReward} 
BLIP & 65.1 &70.2& 41.7 & 59.0\\ 
\cellcolor{myBlue!60}\textbf{HPSv2} 
  ViT-H/14 & 67.9 & 56.4& 36.4 & 53.6 \\ 
\cellcolor{myGreen!50}\textbf{PAC-S}  
ViT-L/14 & 65.8 & \underline{71.2} & \underline{48.0} & \underline{61.6} \\ 
\cellcolor{myRed!50}\textbf{LIQE
} ViT-B/32 & \underline{71.0} & 60.1 & 18.8 &  49.9\\
\cellcolor{myRed!50}\textbf{C2S$^{\clubsuit}$
} mOwl-2 & 61.2 &  - & - & -\\ 
\toprule

\multicolumn{5}{l}{\textbf{Our models}}\\
\toprule
\textbf{UniSim} ViT-B/32 & \textbf{72.9} & 61.9 & 34.2 & 56.3 \\ 
\textbf{UniSim} ViT-L/14 & 67.6 & 53.7 & 25.1 & 48.8 \\ 
\textbf{UniSim}$^{\clubsuit}$ LL-N-0.5B & 64.8 & 24.2 & - & - \\

\bottomrule
\end{tabular}}
\caption{\textbf{Evaluation on the \textit{OOD Generalization Tasks} of \unisimbnc.} We extend the evaluation in Table~\ref{tab:unisim-result} to additional tasks (average accuracy over datasets is reported). LMMs cannot be applied to retrieval tasks. 
Our analysis reveals that the average performance on these unseen tasks (last column) is lower for both specialized perceptual models and our multi-task models compared to the general-purpose baselines.}
\vspace{-3mm}
\label{tab:unisim-result-unseen}
\end{table}

\subsection{Evaluation on \textbf{\gentasks}}
\label{sec:exp_unseen_tasks}

Table~\ref{tab:unisim-result-unseen} reports 
the results on the \gentasks for the models from Table~\ref{tab:unisim-result} (average accuracy over datasets is shown, detailed results 
in App.~\ref{sec:additional_experiments}).
The average performance on these unseen tasks (last column) is lower for perceptual models (both specialized and  multi-task) compared to the general-purpose baselines. 
This aligns with the previous observation that training on a subset of perceptual similarity tasks does not improve performance on unseen tasks.

However, for perceptual attributes assessment (\npaa), specialized models often achieve accuracy close to or slightly exceeding that of the baselines. For example, both CLIP-based UniSim models outperform the original CLIP models from which they are fine-tuned.
We hypothesize that \npaa is a near-OOD task, more similar to the training tasks than odd-one-out or retrieval (far OOD), making multi-task training slightly beneficial in this context.

Finally, unlike for the core tasks (Table~\ref{tab:unisim-result}), the performance of LMMs is generally worse than 
with CLIP models. Notably, the accuracy of the LLM-based UniSim drops significantly on \nooo, likely due to the task structure which differs from the 2AFC training data.
In contrast, CLIP models are unaffected since each input item is encoded independently, regardless of the task.

\begin{table}[t!]
\centering \small
\tabcolsep=1.85pt
\extrarowheight=1.5pt
\myindent=2mm
\newl=7mm
\newcommand{\myangle}{90}
\definecolor{myTableGray}{rgb}{.7, .7, .7}
\definecolor{myTableRed}{rgb}{1., .7, .4}

\scalebox{.84}{
\begin{tabular}{
L{28mm}||
*{3}{C{\newl}}|
*{2}{C{\newl}}
>{\columncolor{myBlue!20!myGray}}C{\newl}|
>{\columncolor{myBlue}}C{\newl}||
>{\columncolor{myGreen!60}}C{\newl}
}
\toprule

 &\multicolumn{7}{c||}{\cellcolor{myBlue} \textbf{\ittwoafc}}  & 
 \cellcolor{myGreen} 
 \\
 
 &\multicolumn{3}{c|}{}  & \multicolumn{3}{c|}{\cellcolor{myBlue!20!myGray} \textbf{unseen}} & & \cellcolor{myGreen} 
 \\

\textbf{\ittwoafc Train set} &
\rotatebox{\myangle}{\hqedit
}  &
\rotatebox{\myangle}{\hpd
} &
\rotatebox{\myangle}{\magicbrush
} &
\rotatebox{\myangle}{\imgreward} &
\rotatebox{\myangle}{\agiqa} &
\rotatebox{\myangle}{\textbf{average}} &
\rotatebox{\myangle}{\textbf{average (all)}} &
\cellcolor{myGreen} \rotatebox{\myangle}{\textbf{\ntexttwoafc}} 
\\
\toprule

\multicolumn{5}{l}{\textbf{UniSim} ViT-B/32}\\
\toprule
%
\hqedit & \textcolor{myTableRed}{85.3} & 71.9 & 62.0 & 67.7 & 69.5 & 68.6 & 71.3 & 89.9 
\\
\hspace{\myindent} + \hpd & \textcolor{myTableRed}{83.6} & \textcolor{myTableRed}{74.1} & 64.2 & \textbf{71.1} & 71.4 & \textbf{71.3} & 72.9 & \textbf{90.4} 
\\
\hspace{2\myindent} + \magicbrush & \textcolor{myTableRed}{84.1} & \textcolor{myTableRed}{74.5} & \textcolor{myTableRed}{78.1} & 70.4 & \textbf{71.7} & 71.1 & \textbf{75.8} & 90.3 
\\
\midrule
\addlinespace[2mm]

\multicolumn{5}{l}{\textbf{UniSim} ViT-L/14}\\
\toprule
\hqedit & \textcolor{myTableRed}{87.9} & 78.7 & 50.9 & 62.6 & 69.0 & 65.8 & 69.8 & 92.5 
\\
\hspace{\myindent} + \hpd & \textcolor{myTableRed}{84.7} & \textcolor{myTableRed}{82.1} & 54.3 & \textbf{71.1} & 70.7 & \textbf{70.9} & 72.6 & 92.3 
\\
\hspace{2\myindent} + \magicbrush & \textcolor{myTableRed}{86.0} & \textcolor{myTableRed}{82.3} & \textcolor{myTableRed}{91.8} & 69.4 & \textbf{71.3} & 70.4 & \textbf{80.2} & \textbf{92.9} 
\\

\bottomrule
\end{tabular}}
\caption{\textbf{Ablation study on the \nittwoafc training data.} When training UniSim we use three \nittwoafc datasets (\hqedit, \hpd, \magicbrush): then we study how using either just one or two of them influences the intra-task generalization (\imgreward, \agiqa) and performance on \ntexttwoafc.
The accuracy on training datasets is in \textcolor{myTableRed}{orange}.
Our findings suggest that incorporating additional \nittwoafc datasets enhances generalization to both unseen datasets and \ntexttwoafc tasks.
}
\label{tab:ablation-data-v2}
\end{table}

\section{Additional Analyses} \label{sec:ablations}

\textbf{Ablation study on the \nittwoafc training data.}
We study here the effect of varying the number of datasets used for training UniSim models.
In particular, we focus on  \nittwoafc, and report in Table~\ref{tab:ablation-data-v2} the results when fine-tuning CLIP models with various configurations.
The default UniSim training uses three datasets (\hqedit, \hpd, \magicbrush), and we test using either just one (\hqedit) or two (\hqedit + \hpd) of them (the training datasets for the other tasks are unchanged).
We find that using two or three datasets (noting that \magicbrush is relatively small, thus has a limited impact) improves intra-task generalization, as observed on \imgreward and \agiqa. Additionally, this setup also enhances performance on a different yet related task, \ntexttwoafc, indicating that jointly training on multiple perceptual tasks can be mutually beneficial.
\\

\begin{figure}[t]
    \centering
    \includegraphics[width=0.47\textwidth]{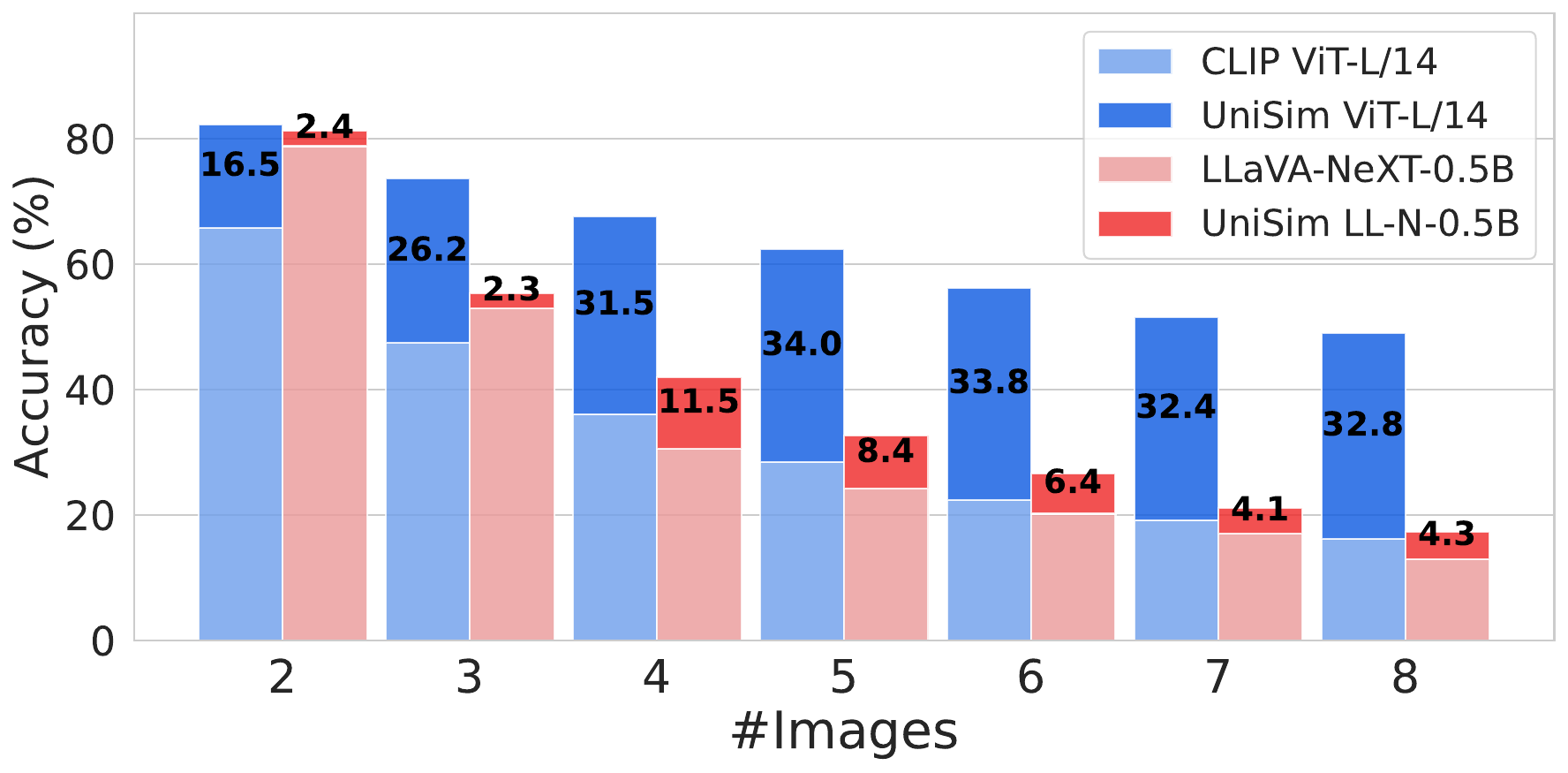}
    \vspace{-2mm}
    \caption{
    \textbf{Increasing the alternatives in Image-to-Text Alignment task.} We report accuracy as the number of alternative images increases in the \nittwoafc (\hpd dataset).
    Both UniSim models preserve higher accuracy than the respective baselines (the gap is highlighted in the plot) as the number of alternatives grows.
    Notably, our encoder-based UniSim ViT-L/14 significantly outperforms the other metrics, including LMM-based UniSim.
    }
    \label{fig:it-nafc}
\end{figure}

\noindent\textbf{From 2AFC to \textit{N}AFC in Image-to-Text Alignment task.}
Next, we analyze the effect of increasing, during the inference, the number of alternative images in \nittwoafc (\hpd dataset) from 2 to \textit{N} (we recall the core tasks in \unisimbnc are 2AFC).
In Fig.~\ref{fig:it-nafc}, we show the accuracy of CLIP and LLaVA-based UniSim, and the corresponding baselines, for $N=2, \ldots, 8$.
Both UniSim models outperform their base models; notably, the CLIP-based UniSim maintains nearly 50\% accuracy at $N=8$, which is three times higher than the base model. Finally, CLIP models significantly outperform LMMs, highlighting some of the current limitations of LMM-based approaches.

\section{Conclusion}
%
Human perception of similarity across modalities is complex and challenging to capture with automated metrics. To advance towards a comprehensive multi-task perceptual model, we introduce \unisimbnc, a benchmark comprising core multi-modal 2AFC perceptual tasks and a diverse set of tasks designed to evaluate out-of-distribution generalization.
Our evaluation on \unisimbnc reveals that single-task specialized metrics often underperform general-purpose models (e.g., CLIP) on unseen tasks. While prior work \cite{sundaram2024does} suggests that aligning multi-modal models with human perception can benefit certain downstream applications, our experiments reveal a more nuanced picture: perceptual metric learning often overfits to the training task, degrading performance on tasks with slightly different formulations, even when closely related to human similarity perception. 
Also, the limited intra-task generalization observed raises questions about the extent to which these models align with human perception in a broad sense.
These insights highlight the need for more robust multi-modal similarity metrics. Our multi-task UniSim models represent a step 
towards this, providing a foundation for future research aimed at capturing more comprehensively the human perception of similarity.
\\

\noindent \textbf{Limitations.}
As discussed above, our collection of tasks and datasets is necessarily limited: for this reason, we formulate \unisimbnc as an open-ended benchmark, which can be iteratively enriched by new additions.
Moreover, we have tested only a small amount of models as a base for our UniSim metrics, as well as training configurations, potentially missing more suitable architectures.

\subsection*{Acknowledgments}
This paper is supported in part by the Army Research Office under grant number W911NF-21-1-0155 and by the New York University Abu Dhabi (NYUAD) Center for Artificial Intelligence and Robotics, funded by Tamkeen under the NYUAD Research Institute Award CG010. Additional support was provided by the NYU IT High Performance Computing resources, services, and staff expertise.
We further acknowledge support from the Swiss National Science Foundation (grant number 212111).

{
    \small
    \bibliographystyle{ieeenat_fullname}
    \bibliography{literatur}

\begin{thebibliography}{58}
\providecommand{\natexlab}[1]{#1}
\providecommand{\url}[1]{\texttt{#1}}
\expandafter\ifx\csname urlstyle\endcsname\relax
  \providecommand{\doi}[1]{doi: #1}\else
  \providecommand{\doi}{doi: \begingroup \urlstyle{rm}\Url}\fi

\bibitem[Achiam et~al.(2023)Achiam, Adler, Agarwal, Ahmad, Akkaya, Aleman, Almeida, Altenschmidt, Altman, Anadkat, et~al.]{achiam2023gpt}
Josh Achiam, Steven Adler, Sandhini Agarwal, Lama Ahmad, Ilge Akkaya, Florencia~Leoni Aleman, Diogo Almeida, Janko Altenschmidt, Sam Altman, Shyamal Anadkat, et~al.
\newblock Gpt-4 technical report.
\newblock \emph{arXiv preprint arXiv:2303.08774}, 2023.

\bibitem[Alabdulmohsin et~al.(2024)Alabdulmohsin, Zhai, Kolesnikov, and Beyer]{alabdulmohsin2024getting}
Ibrahim~M Alabdulmohsin, Xiaohua Zhai, Alexander Kolesnikov, and Lucas Beyer.
\newblock Getting vit in shape: Scaling laws for compute-optimal model design.
\newblock \emph{Advances in Neural Information Processing Systems}, 36, 2024.

\bibitem[Bianco et~al.(2023)Bianco, Celona, Donzella, and Napoletano]{bianco2023improving}
Simone Bianco, Luigi Celona, Marco Donzella, and Paolo Napoletano.
\newblock Improving image captioning descriptiveness by ranking and llm-based fusion.
\newblock \emph{arXiv preprint arXiv:2306.11593}, 2023.

\bibitem[Cai et~al.(2018)Cai, Gu, and Zhang]{cai2018learning}
Jianrui Cai, Shuhang Gu, and Lei Zhang.
\newblock Learning a deep single image contrast enhancer from multi-exposure images.
\newblock \emph{IEEE Transactions on Image Processing}, 27\penalty0 (4):\penalty0 2049--2062, 2018.

\bibitem[Caron et~al.(2021)Caron, Touvron, Misra, Jégou, Mairal, Bojanowski, and Joulin]{caron2021emerging}
Mathilde Caron, Hugo Touvron, Ishan Misra, Hervé Jégou, Julien Mairal, Piotr Bojanowski, and Armand Joulin.
\newblock Emerging properties in self-supervised vision transformers.
\newblock In \emph{ICCV}, 2021.

\bibitem[Cherti et~al.(2023)Cherti, Beaumont, Wightman, Wortsman, Ilharco, Gordon, Schuhmann, Schmidt, and Jitsev]{cherti2023openclip}
Mehdi Cherti, Romain Beaumont, Ross Wightman, Mitchell Wortsman, Gabriel Ilharco, Cade Gordon, Christoph Schuhmann, Ludwig Schmidt, and Jenia Jitsev.
\newblock Reproducible scaling laws for contrastive language-image learning.
\newblock In \emph{CVPR}, 2023.

\bibitem[Croce et~al.(2024)Croce, Schlarmann, Singh, and Hein]{croce2024adversarially}
Francesco Croce, Christian Schlarmann, Naman~Deep Singh, and Matthias Hein.
\newblock Adversarially robust clip models induce better (robust) perceptual metrics.
\newblock In \emph{ICML 2024 Workshop on Foundation Models in the Wild}, 2024.

\bibitem[Fu et~al.(2023)Fu, Tamir, Sundaram, Chai, Zhang, Dekel, and Isola]{fu2023learning}
Stephanie Fu, Netanel Tamir, Shobhita Sundaram, Lucy Chai, Richard Zhang, Tali Dekel, and Phillip Isola.
\newblock Dreamsim: Learning new dimensions of human visual similarity using synthetic data.
\newblock In \emph{NeurIPS}, 2023.

\bibitem[Fu et~al.(2024)Fu, Hu, Li, Feng, Wang, Lin, Roth, Smith, Ma, and Krishna]{fu2024blink}
Xingyu Fu, Yushi Hu, Bangzheng Li, Yu Feng, Haoyu Wang, Xudong Lin, Dan Roth, Noah~A Smith, Wei-Chiu Ma, and Ranjay Krishna.
\newblock Blink: Multimodal large language models can see but not perceive.
\newblock \emph{arXiv preprint arXiv:2404.12390}, 2024.

\bibitem[Ghazanfari et~al.(2023)Ghazanfari, Garg, Krishnamurthy, Khorrami, and Araujo]{ghazanfari2023rlpips}
Sara Ghazanfari, Siddharth Garg, Prashanth Krishnamurthy, Farshad Khorrami, and Alexandre Araujo.
\newblock R-{LPIPS}: An adversarially robust perceptual similarity metric.
\newblock In \emph{ICML Workshop on New Frontiers in Adversarial Machine Learning}, 2023.

\bibitem[Ghazanfari et~al.(2024{\natexlab{a}})Ghazanfari, Araujo, Krishnamurthy, Garg, and Khorrami]{ghazanfari2024emma}
Sara Ghazanfari, Alexandre Araujo, Prashanth Krishnamurthy, Siddharth Garg, and Farshad Khorrami.
\newblock Emma: Efficient visual alignment in multi-modal llms.
\newblock \emph{arXiv preprint arXiv:2410.02080}, 2024{\natexlab{a}}.

\bibitem[Ghazanfari et~al.(2024{\natexlab{b}})Ghazanfari, Araujo, Krishnamurthy, Khorrami, and Garg]{ghazanfari2023lipsim}
Sara Ghazanfari, Alexandre Araujo, Prashanth Krishnamurthy, Farshad Khorrami, and Siddharth Garg.
\newblock Lipsim: A provably robust perceptual similarity metric.
\newblock In \emph{ICLR}, 2024{\natexlab{b}}.

\bibitem[Goodfellow et~al.(2020)Goodfellow, Pouget-Abadie, Mirza, Xu, Warde-Farley, Ozair, Courville, and Bengio]{goodfellow2020generative}
Ian Goodfellow, Jean Pouget-Abadie, Mehdi Mirza, Bing Xu, David Warde-Farley, Sherjil Ozair, Aaron Courville, and Yoshua Bengio.
\newblock Generative adversarial networks.
\newblock \emph{Communications of the ACM}, 63\penalty0 (11):\penalty0 139--144, 2020.

\bibitem[He et~al.(2022)He, Chen, Xie, Li, Dollár, and Girshick]{he2021masked}
Kaiming He, Xinlei Chen, Saining Xie, Yanghao Li, Piotr Dollár, and Ross Girshick.
\newblock Masked autoencoders are scalable vision learners.
\newblock In \emph{CVPR}, 2022.

\bibitem[Hessel et~al.(2021)Hessel, Holtzman, Forbes, Bras, and Choi]{hessel2021clipscore}
Jack Hessel, Ari Holtzman, Maxwell Forbes, Ronan~Le Bras, and Yejin Choi.
\newblock Clipscore: A reference-free evaluation metric for image captioning.
\newblock In \emph{EMNLP}, 2021.

\bibitem[Ho et~al.(2020)Ho, Jain, and Abbeel]{ho2020denoising}
Jonathan Ho, Ajay Jain, and Pieter Abbeel.
\newblock Denoising diffusion probabilistic models.
\newblock \emph{Advances in neural information processing systems}, 33:\penalty0 6840--6851, 2020.

\bibitem[Hosu et~al.(2020)Hosu, Lin, Sziranyi, and Saupe]{hosu2020koniq10k}
Vlad Hosu, Hanhe Lin, Tamas Sziranyi, and Dietmar Saupe.
\newblock Koniq-10k: An ecologically valid database for deep learning of blind image quality assessment.
\newblock \emph{IEEE Transactions on Image Processing}, 29:\penalty0 4041--4056, 2020.

\bibitem[Hu et~al.(2022)Hu, Shen, Wallis, Allen-Zhu, Li, Wang, Wang, and Chen]{hu2022lora}
Edward~J Hu, Yelong Shen, Phillip Wallis, Zeyuan Allen-Zhu, Yuanzhi Li, Shean Wang, Lu Wang, and Weizhu Chen.
\newblock Lo{RA}: Low-rank adaptation of large language models.
\newblock In \emph{ICLR}, 2022.

\bibitem[Huang et~al.(2024)Huang, Sheng, Yang, Yuan, Duan, Chen, Li, Lin, and Shi]{huang2024aesexpert}
Yipo Huang, Xiangfei Sheng, Zhichao Yang, Quan Yuan, Zhichao Duan, Pengfei Chen, Leida Li, Weisi Lin, and Guangming Shi.
\newblock Aesexpert: Towards multi-modality foundation model for image aesthetics perception.
\newblock In \emph{Proceedings of the 32nd ACM International Conference on Multimedia}, pages 5911--5920, 2024.

\bibitem[Hui et~al.(2024)Hui, Yang, Zhao, Shi, Wang, Wang, Zhou, and Xie]{hui2024hq}
Mude Hui, Siwei Yang, Bingchen Zhao, Yichun Shi, Heng Wang, Peng Wang, Yuyin Zhou, and Cihang Xie.
\newblock Hq-edit: A high-quality dataset for instruction-based image editing.
\newblock \emph{arXiv preprint arXiv:2404.09990}, 2024.

\bibitem[Jiang et~al.(2024)Jiang, He, Zeng, Wei, Ku, Liu, and Chen]{jiang2024mantis}
Dongfu Jiang, Xuan He, Huaye Zeng, Cong Wei, Max Ku, Qian Liu, and Wenhu Chen.
\newblock Mantis: Interleaved multi-image instruction tuning.
\newblock \emph{arXiv preprint arXiv:2405.01483}, 2024.

\bibitem[Jinjin et~al.(2020)Jinjin, Haoming, Haoyu, Xiaoxing, Ren, and Chao]{jinjin2020pipal}
Gu Jinjin, Cai Haoming, Chen Haoyu, Ye Xiaoxing, Jimmy~S Ren, and Dong Chao.
\newblock Pipal: a large-scale image quality assessment dataset for perceptual image restoration.
\newblock In \emph{ECCV}, 2020.

\bibitem[Ku et~al.(2023)Ku, Li, Zhang, Lu, Fu, Zhuang, and Chen]{ku2023imagenhub}
Max Ku, Tianle Li, Kai Zhang, Yujie Lu, Xingyu Fu, Wenwen Zhuang, and Wenhu Chen.
\newblock Imagenhub: Standardizing the evaluation of conditional image generation models.
\newblock \emph{arXiv preprint arXiv:2310.01596}, 2023.

\bibitem[Lee et~al.(2021)Lee, Yoon, Dernoncourt, Bui, and Jung]{lee2021umic}
Hwanhee Lee, Seunghyun Yoon, Franck Dernoncourt, Trung Bui, and Kyomin Jung.
\newblock Umic: An unreferenced metric for image captioning via contrastive learning.
\newblock \emph{arXiv preprint arXiv:2106.14019}, 2021.

\bibitem[Li et~al.(2023{\natexlab{a}})Li, Zhang, Wu, Sun, Min, Liu, Zhai, and Lin]{li2023agiqa}
Chunyi Li, Zicheng Zhang, Haoning Wu, Wei Sun, Xiongkuo Min, Xiaohong Liu, Guangtao Zhai, and Weisi Lin.
\newblock Agiqa-3k: An open database for ai-generated image quality assessment.
\newblock \emph{IEEE Transactions on Circuits and Systems for Video Technology}, 2023{\natexlab{a}}.

\bibitem[Li et~al.(2024{\natexlab{a}})Li, Kou, Gao, Cao, Sun, Zhang, Zhou, Zhang, Zhang, Wu, et~al.]{li2024aigiqa}
Chunyi Li, Tengchuan Kou, Yixuan Gao, Yuqin Cao, Wei Sun, Zicheng Zhang, Yingjie Zhou, Zhichao Zhang, Weixia Zhang, Haoning Wu, et~al.
\newblock Aigiqa-20k: A large database for ai-generated image quality assessment.
\newblock \emph{arXiv preprint arXiv:2404.03407}, 2\penalty0 (3):\penalty0 5, 2024{\natexlab{a}}.

\bibitem[Li et~al.(2024{\natexlab{b}})Li, Zhang, Zhang, Zhang, Li, Li, Ma, and Li]{li2024llava}
Feng Li, Renrui Zhang, Hao Zhang, Yuanhan Zhang, Bo Li, Wei Li, Zejun Ma, and Chunyuan Li.
\newblock Llava-next-interleave: Tackling multi-image, video, and 3d in large multimodal models.
\newblock \emph{arXiv preprint arXiv:2407.07895}, 2024{\natexlab{b}}.

\bibitem[Li et~al.(2023{\natexlab{b}})Li, Li, Savarese, and Hoi]{li2023blip2}
Junnan Li, Dongxu Li, Silvio Savarese, and Steven Hoi.
\newblock Blip-2: Bootstrapping language-image pre-training with frozen image encoders and large language models.
\newblock \emph{arXiv:2301.12597}, 2023{\natexlab{b}}.

\bibitem[Lin et~al.(2019)Lin, Hosu, and Saupe]{hanhe2019kadid10k}
Hanhe Lin, Vlad Hosu, and Dietmar Saupe.
\newblock Kadid-10k: A large-scale artificially distorted iqa database.
\newblock In \emph{2019 Tenth International Conference on Quality of Multimedia Experience (QoMEX)}, 2019.

\bibitem[Lin et~al.(2014)Lin, Maire, Belongie, Hays, Perona, Ramanan, Doll{\'a}r, and Zitnick]{lin2014microsoft}
Tsung-Yi Lin, Michael Maire, Serge Belongie, James Hays, Pietro Perona, Deva Ramanan, Piotr Doll{\'a}r, and C~Lawrence Zitnick.
\newblock Microsoft coco: Common objects in context.
\newblock In \emph{ECCV}, 2014.

\bibitem[Link and Heath(1975)]{Link2AFC_1975}
S.~W. Link and R.~A. Heath.
\newblock A sequential theory of psychological discrimination.
\newblock \emph{Psychometrika}, 40:\penalty0 77--105, 1975.

\bibitem[Liu et~al.(2024)Liu, Li, Li, and Lee]{liu2024improved}
Haotian Liu, Chunyuan Li, Yuheng Li, and Yong~Jae Lee.
\newblock Improved baselines with visual instruction tuning.
\newblock In \emph{Proceedings of the IEEE/CVF Conference on Computer Vision and Pattern Recognition}, pages 26296--26306, 2024.

\bibitem[Liu et~al.(2021)Liu, Gong, Wu, Zhang, Su, and Liu]{liu2021fusedream}
Xingchao Liu, Chengyue Gong, Lemeng Wu, Shujian Zhang, Hao Su, and Qiang Liu.
\newblock Fusedream: Training-free text-to-image generation with improved clip+ gan space optimization.
\newblock \emph{arXiv preprint arXiv:2112.01573}, 2021.

\bibitem[Liu et~al.(1907)Liu, Ott, Goyal, Du, Joshi, Chen, Levy, Lewis, Zettlemoyer, and Stoyanov]{liu1907roberta}
Y Liu, M Ott, N Goyal, J Du, M Joshi, D Chen, O Levy, M Lewis, L Zettlemoyer, and V Stoyanov.
\newblock Roberta: A robustly optimized bert pretraining approach. arxiv [preprint](2019).
\newblock \emph{arXiv preprint arXiv:1907.11692}, 1907.

\bibitem[Muttenthaler et~al.(2023)Muttenthaler, Dippel, Linhardt, Vandermeulen, and Kornblith]{muttenthaler2023human}
Lukas Muttenthaler, Jonas Dippel, Lorenz Linhardt, Robert~A. Vandermeulen, and Simon Kornblith.
\newblock Human alignment of neural network representations.
\newblock In \emph{ICLR}, 2023.

\bibitem[Philbin et~al.(2008)Philbin, Chum, Isard, Sivic, and Zisserman]{oxfordparis}
James Philbin, Ondrej Chum, Michael Isard, Josef Sivic, and Andrew Zisserman.
\newblock Lost in quantization: Improving particular object retrieval in large scale image databases.
\newblock In \emph{CVPR}, 2008.

\bibitem[Podell et~al.(2023)Podell, English, Lacey, Blattmann, Dockhorn, M{\"u}ller, Penna, and Rombach]{podell2023sdxl}
Dustin Podell, Zion English, Kyle Lacey, Andreas Blattmann, Tim Dockhorn, Jonas M{\"u}ller, Joe Penna, and Robin Rombach.
\newblock Sdxl: Improving latent diffusion models for high-resolution image synthesis.
\newblock \emph{arXiv preprint arXiv:2307.01952}, 2023.

\bibitem[Prashnani et~al.(2018)Prashnani, Cai, Mostofi, and Sen]{prashnani2018pieapp}
Ekta Prashnani, Hong Cai, Yasamin Mostofi, and Pradeep Sen.
\newblock Pieapp: Perceptual image-error assessment through pairwise preference.
\newblock In \emph{CVPR}, 2018.

\bibitem[Radford et~al.(2021)Radford, Kim, Hallacy, Ramesh, Goh, Agarwal, Sastry, Askell, Mishkin, Clark, Krueger, and Sutskever]{radford2021clip}
Alec Radford, Jong~Wook Kim, Chris Hallacy, Aditya Ramesh, Gabriel Goh, Sandhini Agarwal, Girish Sastry, Amanda Askell, Pamela Mishkin, Jack Clark, Gretchen Krueger, and Ilya Sutskever.
\newblock Learning transferable visual models from natural language supervision.
\newblock In \emph{{ICML}}, 2021.

\bibitem[Sarto et~al.(2023)Sarto, Barraco, Cornia, Baraldi, and Cucchiara]{sarto2023positive}
Sara Sarto, Manuele Barraco, Marcella Cornia, Lorenzo Baraldi, and Rita Cucchiara.
\newblock Positive-augmented contrastive learning for image and video captioning evaluation.
\newblock In \emph{Proceedings of the IEEE/CVF conference on computer vision and pattern recognition}, pages 6914--6924, 2023.

\bibitem[Sundaram et~al.(2024)Sundaram, Fu, Muttenthaler, Tamir, Chai, Kornblith, Darrell, and Isola]{sundaram2024does}
Shobhita Sundaram, Stephanie Fu, Lukas Muttenthaler, Netanel~Y Tamir, Lucy Chai, Simon Kornblith, Trevor Darrell, and Phillip Isola.
\newblock When does perceptual alignment benefit vision representations?
\newblock \emph{arXiv preprint arXiv:2410.10817}, 2024.

\bibitem[Team(2024)]{team2024chameleon}
Chameleon Team.
\newblock Chameleon: Mixed-modal early-fusion foundation models.
\newblock \emph{arXiv preprint arXiv:2405.09818}, 2024.

\bibitem[Wada et~al.(2024)Wada, Kaneda, Saito, and Sugiura]{wada2024polos}
Yuiga Wada, Kanta Kaneda, Daichi Saito, and Komei Sugiura.
\newblock Polos: Multimodal metric learning from human feedback for image captioning.
\newblock In \emph{Proceedings of the IEEE/CVF Conference on Computer Vision and Pattern Recognition}, pages 13559--13568, 2024.

\bibitem[Wang et~al.(2024)Wang, Fu, Huang, Li, Liu, Liu, Ma, Xu, Zhou, Zhang, et~al.]{wang2024muirbench}
Fei Wang, Xingyu Fu, James~Y Huang, Zekun Li, Qin Liu, Xiaogeng Liu, Mingyu~Derek Ma, Nan Xu, Wenxuan Zhou, Kai Zhang, et~al.
\newblock Muirbench: A comprehensive benchmark for robust multi-image understanding.
\newblock \emph{arXiv preprint arXiv:2406.09411}, 2024.

\bibitem[Wang et~al.(2023)Wang, Chan, and Loy]{wang2023exploring}
Jianyi Wang, Kelvin~CK Chan, and Chen~Change Loy.
\newblock Exploring clip for assessing the look and feel of images.
\newblock In \emph{AAAI}, 2023.

\bibitem[Wu et~al.(2023{\natexlab{a}})Wu, Zhang, Zhang, Chen, Liao, Wang, Li, Sun, Yan, Zhai, et~al.]{wu2023q}
Haoning Wu, Zicheng Zhang, Erli Zhang, Chaofeng Chen, Liang Liao, Annan Wang, Chunyi Li, Wenxiu Sun, Qiong Yan, Guangtao Zhai, et~al.
\newblock Q-bench: A benchmark for general-purpose foundation models on low-level vision.
\newblock \emph{arXiv preprint arXiv:2309.14181}, 2023{\natexlab{a}}.

\bibitem[Wu et~al.(2024)Wu, Zhu, Zhang, Zhang, Chen, Liao, Li, Wang, Sun, Yan, et~al.]{wu2024towards}
Haoning Wu, Hanwei Zhu, Zicheng Zhang, Erli Zhang, Chaofeng Chen, Liang Liao, Chunyi Li, Annan Wang, Wenxiu Sun, Qiong Yan, et~al.
\newblock Towards open-ended visual quality comparison.
\newblock \emph{arXiv preprint arXiv:2402.16641}, 2024.

\bibitem[Wu et~al.(2023{\natexlab{b}})Wu, Hao, Sun, Chen, Zhu, Zhao, and Li]{wu2023human}
Xiaoshi Wu, Yiming Hao, Keqiang Sun, Yixiong Chen, Feng Zhu, Rui Zhao, and Hongsheng Li.
\newblock Human preference score v2: A solid benchmark for evaluating human preferences of text-to-image synthesis.
\newblock \emph{arXiv preprint arXiv:2306.09341}, 2023{\natexlab{b}}.

\bibitem[Xu et~al.(2023)Xu, Liu, Wu, Tong, Li, Ding, Tang, and Dong]{xu2023imagereward}
Jiazheng Xu, Xiao Liu, Yuchen Wu, Yuxuan Tong, Qinkai Li, Ming Ding, Jie Tang, and Yuxiao Dong.
\newblock Imagereward: Learning and evaluating human preferences for text-to-image generation.
\newblock In \emph{NeurIPS}, 2023.

\bibitem[Ye et~al.(2024)Ye, Xu, Ye, Yan, Hu, Liu, Qian, Zhang, and Huang]{ye2024mplug}
Qinghao Ye, Haiyang Xu, Jiabo Ye, Ming Yan, Anwen Hu, Haowei Liu, Qi Qian, Ji Zhang, and Fei Huang.
\newblock mplug-owl2: Revolutionizing multi-modal large language model with modality collaboration.
\newblock In \emph{CVPR}, 2024.

\bibitem[Zhai et~al.(2023)Zhai, Mustafa, Kolesnikov, and Beyer]{zhai2023sigmoid}
Xiaohua Zhai, Basil Mustafa, Alexander Kolesnikov, and Lucas Beyer.
\newblock Sigmoid loss for language image pre-training.
\newblock \emph{arXiv preprint arXiv:2303.15343}, 2023.

\bibitem[Zhang et~al.(2023{\natexlab{a}})Zhang, Zhang, Zhang, and Kweon]{zhang2023text}
Chenshuang Zhang, Chaoning Zhang, Mengchun Zhang, and In~So Kweon.
\newblock Text-to-image diffusion models in generative ai: A survey.
\newblock \emph{arXiv preprint arXiv:2303.07909}, 2023{\natexlab{a}}.

\bibitem[Zhang et~al.(2024{\natexlab{a}})Zhang, Mo, Chen, Sun, and Su]{zhang2024magicbrush}
Kai Zhang, Lingbo Mo, Wenhu Chen, Huan Sun, and Yu Su.
\newblock Magicbrush: A manually annotated dataset for instruction-guided image editing.
\newblock \emph{Advances in Neural Information Processing Systems}, 36, 2024{\natexlab{a}}.

\bibitem[Zhang et~al.(2018)Zhang, Isola, Efros, Shechtman, and Wang]{zhang2018unreasonable}
Richard Zhang, Phillip Isola, Alexei~A. Efros, Eli Shechtman, and Oliver Wang.
\newblock The unreasonable effectiveness of deep features as a perceptual metric.
\newblock In \emph{CVPR}, 2018.

\bibitem[Zhang et~al.(2023{\natexlab{b}})Zhang, Zhai, Wei, Yang, and Ma]{zhang2023liqe}
Weixia Zhang, Guangtao Zhai, Ying Wei, Xiaokang Yang, and Kede Ma.
\newblock Blind image quality assessment via vision-language correspondence: A multitask learning perspective.
\newblock In \emph{CVPR}, 2023{\natexlab{b}}.

\bibitem[Zhang et~al.(2024{\natexlab{b}})Zhang, Wu, Li, Zhou, Sun, Min, Chen, Liu, Lin, and Zhai]{zhang2024bench}
Zicheng Zhang, Haoning Wu, Chunyi Li, Yingjie Zhou, Wei Sun, Xiongkuo Min, Zijian Chen, Xiaohong Liu, Weisi Lin, and Guangtao Zhai.
\newblock A-bench: Are lmms masters at evaluating ai-generated images?
\newblock \emph{arXiv preprint arXiv:2406.03070}, 2024{\natexlab{b}}.

\bibitem[Zhu et~al.(2024{\natexlab{a}})Zhu, Sui, Chen, Liu, Chen, Fang, and Wang]{zhu20242afc}
Hanwei Zhu, Xiangjie Sui, Baoliang Chen, Xuelin Liu, Peilin Chen, Yuming Fang, and Shiqi Wang.
\newblock 2afc prompting of large multimodal models for image quality assessment.
\newblock \emph{arXiv preprint arXiv:2402.01162}, 2024{\natexlab{a}}.

\bibitem[Zhu et~al.(2024{\natexlab{b}})Zhu, Wu, Li, Zhang, Chen, Zhu, Fang, Zhai, Lin, and Wang]{zhu2024adaptive}
Hanwei Zhu, Haoning Wu, Yixuan Li, Zicheng Zhang, Baoliang Chen, Lingyu Zhu, Yuming Fang, Guangtao Zhai, Weisi Lin, and Shiqi Wang.
\newblock Adaptive image quality assessment via teaching large multimodal model to compare.
\newblock \emph{arXiv preprint arXiv:2405.19298}, 2024{\natexlab{b}}.

\end{thebibliography}
}

\clearpage

\appendix

\section{Extended Related Work} \label{sec:extended_related_work}

\paragraph{Image-to-Image Similarity Metrics.} Recent perceptual metrics have increasingly leveraged deep neural networks to produce data representations, enabling comparisons in the embedding space through measures such as $\ell_{p}$-norms and cosine similarity~\cite{zhang2018unreasonable, liu2021fusedream, croce2024adversarially}. For image-to-image comparisons, earlier approaches~\cite{zhang2018unreasonable, ghazanfari2023rlpips} utilized the CNN backbones of image classifiers as vision encoders. In contrast, more recent methods~\cite{liu2021fusedream, croce2024adversarially} exploit modern vision foundation models~\cite{radford2021clip, cherti2023openclip, caron2021emerging}, which are trained on vast datasets containing hundreds of millions to billions of samples, to extract highly generalizable visual representations. Additionally, alternative backbones have been explored for visual representation, such as LipSim~\cite{ghazanfari2023lipsim}, which employs Lipschitz networks to enhance robustness against adversarial attacks, and MAE~\cite{he2021masked}, which leverages autoencoders to generate representations.

\paragraph{Image-to-Text Alignment.} With the rise of generative models capable of producing images from textual prompts, there has been an increasing demand for robust multi-modal metrics that can effectively evaluate the alignment between the input prompt and the generated image. CLIP-score~\cite{hessel2021clipscore} and BLIP-score~\cite{li2023blip2} are strong candidates for this task, as their vision and text encoders are specifically trained to produce representations that are aligned. However, the primary challenge is that the scores generated by these models are not well aligned with human preference. To address this issue, recent metrics~\cite{xu2023imagereward, wu2023human} focus on aligning model evaluations with human preferences. These approaches involve collecting datasets that reflect human judgments by presenting prompts alongside pairs of synthetic images and asking participants to select the image that best aligns with the given prompt. Using this data, ImageReward~\cite{xu2023imagereward} fine-tunes a BLIP model, while HPSv2~\cite{wu2023human} fine-tunes a CLIP model, ensuring their outputs are better aligned with human preferences.

\paragraph{Text-to-Image Alignment.} Evaluating the correctness and comprehensiveness of generated captions for images is crucial in the evaluation of vision-language models. Similar to image-to-text alignment, the CLIP-score~\cite{hessel2021clipscore} is leveraged for this task. However, the CLIP model is suboptimal for evaluation metrics because its training data lacks the richness and descriptiveness necessary for evaluating generated long captions as investigated by~\cite{sarto2023positive}. To address this issue,~\citet{sarto2023positive} leverage contrastive learning with augmented positive samples to improve the alignment between captions and visual content on the CLIP architecture. Moreover, Polos~\cite{wada2024polos} proposes a framework for developing metrics based on human feedback and by leveraging pre-trained CLIP and RoBERTa~\cite{liu1907roberta} as the encoders. Note that Polos is excluded from our evaluations because it requires an additional text reference, beyond the image-caption pair, to effectively assess the alignment between the caption and the image.

\paragraph{Image Quality Assessment (IQA).}

With the increasing demand from applications such as super-resolution, denoising, and generative models, the development of advanced IQA methods has gained significant momentum. In this context, foundation models have emerged as the preferred alternative to traditional techniques. Again, vision-language models like CLIP have been effectively employed to compare the visual representations of an image against text prompts describing quality attributes, such as \texttt{``A high-quality photo.''}. From then new variants of CLIP have been introduced that provide specific setups for training and inference. Recent successful approaches include CLIP-IQA~\cite{wang2023exploring}, which introduces an innovative prompt pairing strategy. This method assesses image quality by utilizing the relative distance between the image and two contrasting prompts: \texttt{``Good photo.''} and \texttt{``Bad photo.''}. Moreover, LIQE~\cite{zhang2023liqe} proposes a framework for training IQA task along with auxiliary tasks such as scene classification and distortion type identification to enhance the model's generalization.
Additionally, LMMs have been employed for IQA. Notably, \citet{zhu2024adaptive, liu1907roberta, wu2024towards} utilize mPLUG-Owl2\cite{ye2024mplug} as their base model, fine-tuning it further on IQA datasets. 
While mPLUG-Owl2 operates as a single-image LMM, our proposed model harnesses the capabilities of multi-image LMMs, which are better suited for perceptual tasks involving multiple images.

\section{Details on UniSim Framework} \label{sec:exp_details}

In this section, we detail first the various components of the UniSim framework starting with an overview of \unisimbnc, then the UniSim training process.

\begin{table}[]
    \centering\small
    \tabcolsep=1.5pt
    \resizebox{0.45\textwidth}{!}{%
    \begin{tabular}{L{20mm} L{30mm} | C{16mm} | C{14mm}}
        Task & Dataset& \makecell{UniSim\\ trains on} & \makecell{Test\\ samples} 
        \\
        \toprule
        \rowcolor{black!10!white} \multicolumn{4}{c}{\textbf{\coretasks}}\\
        \toprule
        \multirow{3}{*}{\imgtwoafc} 
         & \nights \cite{fu2023learning} & \checkmark & 1824 \\
         & \bapps~\cite{zhang2018unreasonable} & \ding{55} & 5K \\
         & \pieapp~\cite{prashnani2018pieapp}& \checkmark & 3314 \\
         \midrule
         
         \multirow{5}{*}{\ittwoafc} & \imgreward~\cite{xu2023imagereward} & \ding{55} &  412 \\
         & \hpd~\cite{wu2023human} & \checkmark & 5K \\
         & \agiqa~\cite{li2023agiqa}& \ding{55} & 5K  \\
         & \magicbrush~\cite{zhang2024magicbrush} & \checkmark & 693 \\
         & \hqedit~\cite{hui2024hq} & \checkmark & 2K\\
         \midrule
         
         \multirow{3}{*}{\texttwoafc} & \coco~\cite{bianco2023improving}  & \ding{55}& 780\\
         & \polaris~\cite{wada2024polos} & \checkmark & 5K \\
         & \hqedit~\cite{hui2024hq} & \checkmark & 2K \\
         \midrule
         
         \multirow{5}{*}{\iqa} & \kadid~\cite{hanhe2019kadid10k} & \checkmark & 5K \\
         & \koniq~\cite{hosu2020koniq10k} & \ding{55} & 5K \\
         & \pieapp~\cite{prashnani2018pieapp} & \checkmark & 5K \\
         & \agiqa~\cite{li2023agiqa} & \ding{55} &  5K \\
         & \pipal~\cite{jinjin2020pipal} & \checkmark & 3025 \\

         \addlinespace[.5mm]
        \rowcolor{black!10!white} \multicolumn{4}{c}{\textbf{\gentasks}}\\
        \toprule
         \multirow{2}{*}{\paa} & \sice~\cite{cai2018learning} & \ding{55} & 2151  \\
         & \koniq~\cite{,hosu2020koniq10k} & \ding{55} & 4 x 5K \\
         \midrule
         
         \multirow{2}{*}{\ooo} & \cifarhooo~\cite{muttenthaler2023human} & \ding{55} & 5K \\
         & \imgnet  & \ding{55} & 5K\\
         \midrule
         
         \multirow{2}{*}{\ir} & \oxford~\cite{oxfordparis} & \ding{55} &  70 \\
         & \paris~\cite{oxfordparis} & \ding{55} &  70 \\
        \midrule
        \multicolumn{3}{l}{\textbf{Total}} & \normalsize{\textbf{88K}}\\
         \bottomrule
    \end{tabular}}
    \caption{\textbf{Composition of UniSim-Bench.}
    We details the datasets used to evaluate each task in our benchmark, as well as whether they are used to train our UniSim models.
    }
    \label{tab:unisim-bench-composition}
\end{table}

\subsection{Perceptual Tasks \& Datasets in \unisimbnc}
\label{sec:details_datasets}

In the following, each paragraph is dedicated to a specific perceptual task covered in \unisimbnc and its associated datasets (also summarized in Table~\ref{tab:unisim-bench-composition}), and complements the descriptions in Sec.~\ref{sec:tasks}.

\paragraph{Image-to-Image Similarity (\imgtwoafc).}
In this task, each data point consists of a triplet $(\vx_\textrm{ref}, \vx_1, \vx_2)$, and one has to decide which of two images $\vx_1, \vx_2$ is most similar to the reference image $\vx_\textrm{ref}$.
The \bapps dataset \cite{zhang2018unreasonable} contains patches of real images perturbed with different corruptions, and compares their similarity to the original images: this was used to tune the LPIPS metric.
A similar approach is used to build \pieapp \cite{prashnani2018pieapp}, where many distortion are applied natural images.
Conversely, \nights \cite{fu2023learning} includes high resolution synthetic images, and aims at capturing similarity in terms of pose, perspective, foreground color, number of items, and object shape.
All datasets contain labels describing the human preference over the alternative images.

\paragraph{Image-to-Text Alignment (\ittwoafc).}
Perceptual metrics are utilized to assess the quality of synthetic images produced by text-to-image generative models~\cite{goodfellow2020generative, ho2020denoising}, evaluating both the overall image quality and the alignment between the provided description and the generated image, ensuring that all relevant details are accurately captured. To achieve this, the \imgreward~\cite{xu2023imagereward} dataset was curated, comprising six synthetic images for each prompt, with a total of 412 
prompts in the test set which are then ranked by experts to capture human preferences for text-to-image generation.
For each prompt, we compare the images with highest and lowest rank, to have confident ground-truth labels.
Additionally, the \hpd dataset~\cite{wu2023human} was introduced as a large-scale collection aimed at capturing human preferences across a wide variety of image sources. It comprises 798,090 human preference annotations across 433,760 image pairs, making it one of the largest datasets of its kind. 
The test set samples consist of a prompt, multiple images, and ranks indicating the alignment of each image with the prompt. Following the \nittwoafc setting, two images are randomly selected, and the label is assigned based on their respective rankings. Another dataset utilized in this area is called \agiqa~\cite{li2023agiqa}, designed to evaluate the subjective quality of AI-generated images. It provides subjective scores for two key aspects: perceptual quality, which assesses the overall visual appeal and realism of the images, and text-to-image alignment, which evaluates how well the generated image corresponds to the given textual description. 
For our benchmark, we first filter out images with low perceptual quality scores. Then, two images are randomly selected and labeled based on the alignment score to form a \nittwoafc sample.
The area of instruction-guided image editing features datasets in a structured format, comprising source images, textual instructions, and target images. These datasets naturally align with the \nittwoafc task, as basically, the instruction is a description of the target image. Consequently, we have utilized the \magicbrush~\cite{zhang2024magicbrush} and \hqedit~\cite{hui2024hq} from this literature to capitalize on their detailed annotations and structured triplets.
\hqedit provides textual descriptions for both the source and target images. Consequently, each sample effectively becomes two distinct samples by utilizing one description at a time and swapping the label accordingly.
\paragraph{Text-to-Image Alignment (\texttwoafc).} The majority of the literature on perceptual metrics has concentrated on evaluating the quality and alignment of synthetic images produced by generative models. However, the reverse task—where an image serves as the input and text is generated as the output—is equally significant. Assessing the quality and specificity of generated captions is essential for ensuring accurate and meaningful text generation. To address this gap, we incorporate the \ntexttwoafc task, as one of the important tasks for multi-modal perceptual metrics. For this task, we utilize three datasets including \polaris~\cite{wada2024polos}, \coco~\cite{bianco2023improving} and \hqedit~\cite{hui2024hq}. The \polaris dataset consists of 131,020 generated captions and 262,040 reference captions, with human evaluations gathered from 550 participants. Each sample includes an image, a reference caption, and generated captions that received a score of 0.5 or lower. The \coco~\cite{bianco2023improving} benchmark utilizes the MS-COCO~\cite{lin2014microsoft} dataset and generates multiple captions for each image using advanced captioning models and by fusing the top two captions richer, more descriptive captions are created. We utilize 1,000 samples that have human annotations and prune the ones with negative votes and by pairing them with five original captions of MS-COCO data, we create a total of 780 paired samples for evaluations. Finally, the \hqedit dataset, introduced in the previous section, is particularly well-suited for this task as it provides detailed descriptions for both source and target images. Each sample in the \ntexttwoafc task comprises either a combination of the source image, source description, and target description or the target image paired with the source and target descriptions.

\paragraph{Image Quality Assessment (\iqa).}
This is an established task where one has to determine which of two images $\vx_1, \vx_2$ has higher quality.
While there exist works focusing on no-reference quality assessment, i.e. an absolute score, we here restrict our evaluation to pairwise comparison.
The \kadid dataset~\cite{hanhe2019kadid10k} contains artificially corrupted images with varying levels of severity. Each corrupted image corresponds to a specific reference image. To generate a single sample for \niqa, we randomly select an image from the dataset and pair it with another image that represents the next severity level of corruption. Similarly, the \koniq dataset consists of a pool of images with authentic distortions, from which two images are randomly selected to form a sample
Additionally, the \pieapp dataset can be leveraged by comparing original images with their corresponding corrupted versions. As previously discussed, the \agiqa dataset provides both a perceptual quality score and an image-text alignment score, making it an excellent resource for evaluating the Image Quality Assessment (\niqa) 
by utilizing the the perceptual quality score.
Another dataset for \niqa is the \pipal~\cite{jinjin2020pipal} dataset comprising 29,000 images, including 250 high-quality reference images, each subjected to 116 types of distortions. To ensure reliable subjective quality scores, the dataset includes over 1.13 million human judgments for annotation.

\paragraph{Perceptual Attributes Assessment (\paa).} This task refers to the evaluation of specific visual characteristics or qualities of the image that directly influence how it is perceived by humans. These attributes are subjective and involve measuring various aspects of the image's appearance that contribute to its overall visual quality. More specifically the perceptual attributes included in our work consist of  brightness (the perceived level of light or luminance in the image), colorfulness (the intensity or vibrancy of the colors in the image), contrast (the degree of difference between the darkest and lightest parts of the image)
and sharpness (the clarity or focus of details in the image). For brightness evaluation, we utilize the \sice~\cite{cai2018learning} dataset, while for other attributes, including brightness, we leverage the \koniq~\cite{hosu2020koniq10k} dataset. More specifically, both datasets contain a pool of images with varying levels of the associated perceptual attribute. To create a sample, two images are randomly selected from the pool, and the label is assigned to the image with the higher perceptual attribute level.

\paragraph{Odd-One-Out (\ooo).}

Given a triplet of images, the task consists in finding the one that does not belong with the others, that is the most dissimilar one.
We use the dataset derived by \cite{muttenthaler2023human} from the coarse \cifarh classes, named \cifarhooo.
Moreover, we follow a similar approach to obtain \imgnet: we create 6 macro-classes (\texttt{aquatic animals}, \texttt{terrestrial animals}, \texttt{clothes}, \texttt{transportations}, \texttt{places}, \texttt{musical instruments}) merging a subset of the \imagenet-1k classes: in this way we get sufficiently semantically separated classes but with enough intra-class diversity so that the tasks is not trivial.
Then, for each triplet we sample two images from a macro-class and one from another, which is the ground-truth odd-one-out image.
We name this dataset \imgnet.

\begin{table}[]
    \centering \small
    \tabcolsep=1.5pt
    \begin{tabular}{L{20mm} L{26mm}|C{16mm} | C{15mm} } 
        \toprule
        \textbf{Task} & \textbf{Dataset} & \textbf{Type} & \makecell{\textbf{Train}\\ \textbf{samples}} \\ 
        \toprule
        \multirow{2}{*}{\textbf{\imgtwoafc}} 
         & \nights \cite{fu2023learning} & Synthetic  & 15.9K \\ 
         & \pieapp~\cite{prashnani2018pieapp} & Realistic & 50.5K \\ 
         \midrule
         \multirow{3}{*}{\textbf{\ittwoafc}} & \hpd~\cite{wu2023human}  & \multirow{3}{*}{Synthetic}  & 645.1K \\ 
         & \magicbrush~\cite{zhang2024magicbrush}  &   & 11.5K \\ 
         & \hqedit~\cite{hui2024hq}  &   & 100K \\
         \midrule
         
         \multirow{2}{*}{\textbf{\texttwoafc}} & \polaris~\cite{wada2024polos}& Realistic  & 245.9K \\ 
         & \hqedit~\cite{hui2024hq}  & Synthetic  & 100K \\
         \midrule
         
         \multirow{3}{*}{\textbf{\iqa}} & \kadid~\cite{hanhe2019kadid10k} & \multirow{3}{*}{Realistic}  & 9.1K \\
         & \pieapp~\cite{prashnani2018pieapp}  &  & 50.5K \\ 
         & \pipal~\cite{jinjin2020pipal} &  & 73.7K \\
         \midrule
        \multicolumn{2}{l}{\textbf{Total}} & & \normalsize{\textbf{1.3M}} \\ 
         \bottomrule

    \end{tabular}
    \caption{\textbf{Overview of UniSim training data.}
    We report the composition of the dataset used for fine-tuning the UniSim models. The number of samples refers to the total contained in the datasets, but might differ from what effectively seen during training by the UniSim models  (for example we balance the number of samples from each task while fine-tuning CLIP).
    }
    \label{tab:unisim-train-data}
\end{table}

\paragraph{Image Retrieval (\ir).}
Perceptual metrics have long been employed to identify the closest matches to a query image within a database of images. In this work we employ the revisited versions of  Oxford and Paris datasets~\cite{oxfordparis}. Both datasets offer three evaluation protocols (easy, medium, hard) to assess performance across varying difficulty levels. \oxford contains around 5,000 images in the retrieval pool, while \paris includes around 6,000 images, and each use 70 query images. For our evaluations, we report the average accuracy on the medium and hard difficulty levels.

\subsection{UniSim Training}

\begin{table*}
    \centering
    \resizebox{\textwidth}{!}{%
    \begin{tabular}{l|c}
    \toprule
         \textbf{Tasks} &  \textbf{Instruction}\\
         \toprule
         \imgtwoafc & \makecell{Answer the following multiple-choice question:\newll Here are three images: \imgtoken \imgtoken \imgtoken. \newll \\ If image 1 is the reference image, which image of the other two is more similar to the reference image? \newll \\ Options: \newll (A) Image 2 \newll (B) Image 3}\\
        \midrule
         \ittwoafc & \makecell{Answer the following question:\newll Here are two images: \imgtoken \imgtoken, \\
         and here is the reference caption: \prompt. which of the two images is more aligned to the reference caption?\newll \\ Options:\newll (A) Image 1 \newll (B) Image 2} \\
         \midrule
         \texttwoafc & \makecell{Answer the following multiple-choice question:\newll Given the reference image: \imgtoken \\ and two captions, caption 1: \{\texttt{caption1}\}, caption 2: \{\texttt{caption2}\} \newll \\ which caption has a better alignment with the reference image? \\ \newll Options:\newll (A) Caption 1\newll (B) Caption 2}\\
         \midrule
         \iqa & \makecell{Answer the following multiple-choice question:\newll Given two images: \imgtoken  \imgtoken \\ which image has a better quality? 
                               \newll Options:\newll (A) Image 1\newll (B) Image 2} \\
         \midrule
         \paa & \makecell{Answer the following multiple-choice question:\newll Given two images: \imgtoken \imgtoken \\ which image is more \{\texttt{perceptual attribute}\}? \newll Options:\newll (A) Image 1 \newll(B) Image 2} \\
         \midrule
         \ooo & \makecell{Answer the following multiple-choice question:\newll Here are three images: \imgtoken \imgtoken  \imgtoken, \\ Which one (A, B, C) is the odd-one-out of the group?\newll \\
                    Options:\newll (A) Image 1\newll(B) Image 2\newll(C) Image 3'} \\
         \bottomrule
    \end{tabular}}
    \caption{\textbf{Instructions employed during inference for each perceptual task.} We detail the prompt used for evaluating the LMMs on the various perceptual tasks.}
    \label{tab:prompts}
\end{table*}

In this section, we present the implementation details of our proposed perceptual metrics, \unisimvlm and \unisimlmm, which are based on
encoder and generative multi-modal models, respectively. 

\paragraph{\unisimvlm: Encoder-based Perceptual Metric.}
For the training of \unisimvlm, we experiment with different versions of CLIP, including \href{https://huggingface.co/laion/CLIP-ViT-B-32-laion2B-s34B-b79K}{ViT-B/32} and \href{https://huggingface.co/openai/clip-vit-large-patch14}{ViT-L/14}, which vary in patch size,  model size and image resolution. For the training data, the datasets presented in Table~\ref{tab:unisim-train-data} are utilized. 
To ensure a balanced number of samples across tasks, we randomly select 400K samples from each task, resulting in a total of 1.6M samples for training. To ensure consistency, a unified training configuration is employed across all versions, including the use of hinge loss with a margin of 0.05, a batch size of 32, only one epoch with a maximum learning rate of $5 \times 10^{-6}$, a weight decay of 0.35, and a warm-up period of 500 steps, following a cosine learning rate schedule.
Moreover, we leverage LoRA (Low-Rank Adaptation)~\cite{hu2022lora} (with rank=16, alpha=32, and dropout=0.2) as employed in the previous works~\cite{liu2021fusedream,croce2024adversarially} to enable efficient fine-tuning while mitigating overfitting.

\begin{table*}[h]
\centering \small
\tabcolsep=1.2pt
\extrarowheight=1.5pt
\myindent=4mm
\newl=7mm
\newcommand{\myangle}{90}
\definecolor{myTableGray}{rgb}{.0, .0, .0}

\resizebox{\textwidth}{!}{%
\begin{tabular}{L{32mm}||
*{3}{C{\newl}}
>{\columncolor{myYellow!50}}C{\newl}|
*{5}{C{\newl}}
>{\columncolor{myBlue!50}}C{\newl}|
*{3}{C{\newl}}
>{\columncolor{myGreen!60}}C{\newl}|
*{5}{C{\newl}}
>{\columncolor{myRed!50}}C{\newl}||
>{\columncolor{myGray!50}}C{\newl}
}
\toprule

& \multicolumn{4}{c}{\cellcolor{myYellow} \textbf{\imgtwoafc}}  & \multicolumn{6}{c}{\cellcolor{myBlue} \textbf{\ittwoafc}}  & \multicolumn{4}{c}{\cellcolor{myGreen} \textbf{\texttwoafc}}  & \multicolumn{6}{c}{\cellcolor{myRed} \textbf{\iqa}}  & \textbf{Avg} \\

Models  & 
\rotatebox{\myangle}{\nights$^{(1,\dagger)}$} &
\rotatebox{\myangle}{\bapps} &
\rotatebox{\myangle}{\pieapp$^{(\dagger)}$}  & 
\rotatebox{\myangle}{\textbf{average}} &
\rotatebox{\myangle}{\imgreward$^{(3)}$} &
\rotatebox{\myangle}{\hpd$^{(4, \dagger)}$} &
\rotatebox{\myangle}{\agiqa} &
\rotatebox{\myangle}{\magicbrush$^{(\dagger)}$} &
\rotatebox{\myangle}{\hqedit$^{(\dagger)}$}  & 
\rotatebox{\myangle}{\textbf{average}} &
\rotatebox{\myangle}{\coco} &
\rotatebox{\myangle}{\polaris$^{(\dagger)}$} &
\rotatebox{\myangle}{\hqedit$^{(\dagger)}$}  & 
\rotatebox{\myangle}{\textbf{average}} &
\rotatebox{\myangle}{\kadid$^{(5, \dagger)}$} &
\rotatebox{\myangle}{\koniq$^{(6)}$} &
\rotatebox{\myangle}{\pieapp$^{(\dagger)}$} &
\rotatebox{\myangle}{\agiqa} &
\rotatebox{\myangle}{\pipal$^{(\dagger)}$}  & 
\rotatebox{\myangle}{\textbf{average}} &
\\
\toprule
\multicolumn{10}{l}{\textbf{General-purpose models}}\\
\toprule

\textbf{CLIP} ViT-B/32 & \textcolor{myTableGray}{85.1}  & \textcolor{myTableGray}{68.6}  & \textcolor{myTableGray}{80.2}  & 78.0  & \textcolor{myTableGray}{65.8}  & \textcolor{myTableGray}{63.3}  & \textcolor{myTableGray}{66.1}  & \textcolor{myTableGray}{72.4}  & \textcolor{myTableGray}{85.2}  & 70.6 & \textcolor{myTableGray}{61.4}  & \textcolor{myTableGray}{78.9}  & \textcolor{myTableGray}{84.6}  & 75.0 &
\textcolor{myTableGray}{59.8} & \textcolor{myTableGray}{51.8} & \textcolor{myTableGray}{80.5} & \textcolor{myTableGray}{68.3} & \textcolor{myTableGray}{74.4} &	67.0 & 72.6\\

\textbf{CLIP} ViT-L/14 & \textcolor{myTableGray}{81.5} & \textcolor{myTableGray}{64.2} & \textcolor{myTableGray}{76.1} & 73.9  & \textcolor{myTableGray}{63.1} & \textcolor{myTableGray}{65.8} & \textcolor{myTableGray}{62.9} & \textcolor{myTableGray}{78.2} & \textcolor{myTableGray}{84.7} & 70.9  & \textcolor{myTableGray}{75.0} & \textcolor{myTableGray}{82.0} & \textcolor{myTableGray}{83.6} & 80.2  & \textcolor{myTableGray}{84.1} & \textcolor{myTableGray}{69.1} & \textcolor{myTableGray}{90.5} & \textcolor{myTableGray}{77.7} & \textcolor{myTableGray}{88.8} & 82.0  & 76.8 \\

\textbf{CLIP} ViT-H/14  & \textcolor{myTableGray}{84.0} & \textcolor{myTableGray}{69.0} & \textcolor{myTableGray}{76.8} & 76.6  & \textcolor{myTableGray}{63.3} & \textcolor{myTableGray}{65.5} & \textcolor{myTableGray}{65.1} & \textcolor{myTableGray}{76.5} & \textcolor{myTableGray}{86.5} & 71.4  & \textcolor{myTableGray}{66.4} & \textcolor{myTableGray}{81.8} & \textcolor{myTableGray}{85.6} & 77.9  & \textcolor{myTableGray}{67.0} & \textcolor{myTableGray}{61.1} & \textcolor{myTableGray}{72.0} & \textcolor{myTableGray}{65.7} & \textcolor{myTableGray}{67.5} &	66.7  & 73.1\\

\textbf{BLIP} ViT-L/14  & \textcolor{myTableGray}{80.8} & \textcolor{myTableGray}{65.0} & \textcolor{myTableGray}{72.1} & 72.6  & \textcolor{myTableGray}{64.1} & \textcolor{myTableGray}{67.0} & \textcolor{myTableGray}{64.5} & \textcolor{myTableGray}{73.3} & \textcolor{myTableGray}{85.4} & 70.9  & \textcolor{myTableGray}{66.3} & \textcolor{myTableGray}{78.9} & \textcolor{myTableGray}{82.6} & 75.9  & \textcolor{myTableGray}{65.1} & \textcolor{myTableGray}{55.2} & \textcolor{myTableGray}{61.0} & \textcolor{myTableGray}{57.0} & \textcolor{myTableGray}{61.9} & 60.0  & 69.9 \\

\textbf{SigLIP} SoViT/14 
& \textcolor{myTableGray}{82.8} & \textcolor{myTableGray}{66.8} & \textcolor{myTableGray}{78.8} & 76.1  & \textcolor{myTableGray}{63.8} & \textcolor{myTableGray}{69.2} & \textcolor{myTableGray}{65.5} & \textcolor{myTableGray}{75.2} & \textcolor{myTableGray}{79.1} & 70.6  & \textcolor{myTableGray}{66.2} & \textcolor{myTableGray}{82.0} & \textcolor{myTableGray}{76.5} & 74.9  & \textcolor{myTableGray}{57.2} & \textcolor{myTableGray}{55.2} & \textcolor{myTableGray}{62.5} & \textcolor{myTableGray}{59.3} & \textcolor{myTableGray}{61.5} & 59.1 & 70.2 \\

\textbf{LLaVA-NeXT}-0.5B$^{\clubsuit}$  & \textcolor{myTableGray}{58.7} & \textcolor{myTableGray}{52.8} & \textcolor{myTableGray}{63.0} & 58.2  & \textcolor{myTableGray}{61.3} & \textcolor{myTableGray}{76.6} & \textcolor{myTableGray}{65.2} & \textcolor{myTableGray}{64.4} & \textcolor{myTableGray}{75.1} & 68.5  & \textcolor{myTableGray}{53.7} & \textcolor{myTableGray}{71.6} & \textcolor{myTableGray}{57.9} & 61.1  & \textcolor{myTableGray}{53.6} & \textcolor{myTableGray}{52.7} & \textcolor{myTableGray}{55.1} & \textcolor{myTableGray}{57.5} & \textcolor{myTableGray}{50.8} & 53.9  & 60.4 \\

\textbf{LLaVA-NeXT}-7B$^{\clubsuit}$  & \textcolor{myTableGray}{91.3} & \textcolor{myTableGray}{67.0} & \textcolor{myTableGray}{79.9} & 79.4  & \textcolor{myTableGray}{71.5} & \textcolor{myTableGray}{76.1} & \textcolor{myTableGray}{68.5} & \textcolor{myTableGray}{72.7} & \textcolor{myTableGray}{86.5} & 75.1  & \textcolor{myTableGray}{59.6} & \textcolor{myTableGray}{79.4} & \textcolor{myTableGray}{80.0} & 73.0  & \textcolor{myTableGray}{66.1} & \textcolor{myTableGray}{79.2} & \textcolor{myTableGray}{83.6} & \underline{\textcolor{myTableGray}{80.6}} & \textcolor{myTableGray}{80.9} & 78.1  & 76.4 \\

\textbf{Mantis} Idefics-8B$^{\clubsuit}$  & \textcolor{myTableGray}{89.5} & \textcolor{myTableGray}{63.8} & \textcolor{myTableGray}{75.0} & 76.1  & \textcolor{myTableGray}{71.0} & \textcolor{myTableGray}{73.9} & \textcolor{myTableGray}{68.5} & \textcolor{myTableGray}{75.8} & \textcolor{myTableGray}{84.4} & 74.7  & \textcolor{myTableGray}{64.7} & \textcolor{myTableGray}{77.8} & \textcolor{myTableGray}{83.0} & 75.2  & \textcolor{myTableGray}{58.3} & \textcolor{myTableGray}{76.3} & \textcolor{myTableGray}{65.1} & \textcolor{myTableGray}{79.0} & \textcolor{myTableGray}{74.9} & 70.7  & 74.2 \\
\midrule

\multicolumn{21}{l}{\textbf{Specialized models}} \\
\toprule
\cellcolor{myYellow!50}\textbf{DS}$^{(1)}$ ViT-B/32  & \underline{\textcolor{myTableGray}{95.3}} & \textbf{\textcolor{myTableGray}{73.3}} & \underline{\textcolor{myTableGray}{88.5}} & \underline{85.7} & \textcolor{myTableGray}{63.1} & \textcolor{myTableGray}{62.0} & \textcolor{myTableGray}{64.4} & \textcolor{myTableGray}{68.8} & \textcolor{myTableGray}{79.8} & 67.6  & \textcolor{myTableGray}{61.3} & \textcolor{myTableGray}{75.6} & \textcolor{myTableGray}{84.1} & 73.7  & \textcolor{myTableGray}{70.1} & \textcolor{myTableGray}{58.0} & \textcolor{myTableGray}{78.4} & \textcolor{myTableGray}{67.1} & \textcolor{myTableGray}{72.7} & 69.2  & 74.1 \\

\cellcolor{myYellow!50}\textbf{DS}$^{(1)}$ Ensemble  & \textcolor{myTableGray}{\textbf{96.2}} & \textcolor{myTableGray}{\underline{72.5}}
 & \textcolor{myTableGray}{\textbf{89.1}} & \textbf{85.9}  & \textcolor{myTableGray}{-} & \textcolor{myTableGray}{-} & \textcolor{myTableGray}{-} & \textcolor{myTableGray}{-} & \textcolor{myTableGray}{-} & -  & \textcolor{myTableGray}{-} & \textcolor{myTableGray}{-} & \textcolor{myTableGray}{-} & -  & \textcolor{myTableGray}{-} & \textcolor{myTableGray}{-} & \textcolor{myTableGray}{-} & \textcolor{myTableGray}{-} & \textcolor{myTableGray}{-} & -  & -\\

\cellcolor{myBlue!60}\textbf{IR}$^{(3)}$ BLIP 
& \textcolor{myTableGray}{87.1} & \textcolor{myTableGray}{66.1} & \textcolor{myTableGray}{77.6} & 76.9 & \textbf{\textcolor{myTableGray}{74.3}} & \textcolor{myTableGray}{74.5} & \textcolor{myTableGray}{\underline{72.4}} & \textcolor{myTableGray}{74.3} & \textcolor{myTableGray}{83.5} & 75.8  & \textcolor{myTableGray}{54.2} & \textcolor{myTableGray}{72.2} & \textcolor{myTableGray}{85.4} & 70.6  &

\textcolor{myTableGray}{62.3} & \textcolor{myTableGray}{58.0} & \textcolor{myTableGray}{75.1} & \textcolor{myTableGray}{74.8} & \textcolor{myTableGray}{60.1} & 66.1 & 72.3
\\

\cellcolor{myBlue!60}\textbf{HPSv2}$^{(4)}$ ViT-H/14  & \textcolor{myTableGray}{78.5} & \textcolor{myTableGray}{66.7} & \textcolor{myTableGray}{70.8} & 72.0  & \textcolor{myTableGray}{\underline{73.8}} & \textbf{\textcolor{myTableGray}{83.5}} & \textbf{\textcolor{myTableGray}{72.6}} & \textcolor{myTableGray}{74.9} & \textcolor{myTableGray}{81.2} & 77.2  & \textcolor{myTableGray}{68.2} & \textcolor{myTableGray}{78.1} & \textcolor{myTableGray}{81.5} & 75.9  & 
\textcolor{myTableGray}{67.0} & \textcolor{myTableGray}{63.6} & \textcolor{myTableGray}{68.9} & \textcolor{myTableGray}{65.4} & \textcolor{myTableGray}{73.5} & 67.7  & 73.2\\

\cellcolor{myGreen!50}\textbf{PAC-S} ViT-L/14  & \textcolor{myTableGray}{86.9} & \textcolor{myTableGray}{69.1} & \textcolor{myTableGray}{78.1} & 78.0 & \textcolor{myTableGray}{65.0} & \textcolor{myTableGray}{67.0} & \textcolor{myTableGray}{65.8} & \textcolor{myTableGray}{75.6} & \textcolor{myTableGray}{86.9} & 72.1  & \textcolor{myTableGray}{60.5} & \textcolor{myTableGray}{77.6} & \textcolor{myTableGray}{85.6} & 74.6  & \textcolor{myTableGray}{75.0} & \textcolor{myTableGray}{56.5} & \textcolor{myTableGray}{86.1} & \textcolor{myTableGray}{70.0} & \textcolor{myTableGray}{83.2} & 74.2  & 74.7 \\

\cellcolor{myRed!50}\textbf{LIQE$^{(5,6)}$} ViT-B/32   
& \textcolor{myTableGray}{77.9} & \textcolor{myTableGray}{68.7} & \textcolor{myTableGray}{76.6} & 74.4 & \textcolor{myTableGray}{61.9} & \textcolor{myTableGray}{67.3} & \textcolor{myTableGray}{64.1} & \textcolor{myTableGray}{59.9} & \textcolor{myTableGray}{78.3} & 66.3 & \textcolor{myTableGray}{63.5} & \textcolor{myTableGray}{78.2} & \textcolor{myTableGray}{81.0} & 74.2 & 
\textcolor{myTableGray}{92.4} & \textcolor{myTableGray}{\underline{87.9}} & \textcolor{myTableGray}{98.2} & \textcolor{myTableGray}{76.7} & \textcolor{myTableGray}{86.0} & \underline{88.2} & 75.8
\\

\cellcolor{myRed!50}\textbf{C2S$^{\clubsuit}$$^{(5,6)}$} mOwl-2 & \textcolor{myTableGray}{-} & \textcolor{myTableGray}{-} & \textcolor{myTableGray}{-} & -  & \textcolor{myTableGray}{-} & \textcolor{myTableGray}{-} & \textcolor{myTableGray}{-} & \textcolor{myTableGray}{-} & \textcolor{myTableGray}{-} & -  & \textcolor{myTableGray}{-} & \textcolor{myTableGray}{-} & \textcolor{myTableGray}{-} & -  & \textbf{\textcolor{myTableGray}{96.2}} & \textbf{\textcolor{myTableGray}{92.0}} & \textbf{\textcolor{myTableGray}{99.2}} & \textcolor{myTableGray}{76.3} & \textcolor{myTableGray}{87.3} & \textbf{90.2}  & - \\

\midrule

\multicolumn{21}{l}{\textbf{Our models}$^{(\dagger)}$} \\
\toprule
\textbf{UniSim} ViT-B/32  & \textcolor{myTableGray}{87.7} & \textcolor{myTableGray}{69.9} & \textcolor{myTableGray}{84.6} & 80.7  & \textcolor{myTableGray}{70.4} & \textcolor{myTableGray}{74.5} & \textcolor{myTableGray}{71.7} & \textcolor{myTableGray}{78.1} & \textcolor{myTableGray}{84.1} & 75.8  & \textcolor{myTableGray}{\underline{91.2}} & \textcolor{myTableGray}{94.2} & \textcolor{myTableGray}{85.6} & \underline{90.3}  & \textcolor{myTableGray}{89.9} & \textcolor{myTableGray}{72.0} & \textcolor{myTableGray}{93.6} & \textcolor{myTableGray}{77.3} &	\textcolor{myTableGray}{\textbf{93.4}} & 85.3  & 83.0 \\

\textbf{UniSim} ViT-L/14  & \textcolor{myTableGray}{90.7} & \textcolor{myTableGray}{68.1} & \textcolor{myTableGray}{85.0} & 81.3  & \textcolor{myTableGray}{69.4} & \textcolor{myTableGray}{\underline{82.3}} & \textcolor{myTableGray}{71.3} & \textcolor{myTableGray}{91.8} & \textcolor{myTableGray}{86.0} & \underline{80.2}  & \textbf{\textcolor{myTableGray}{94.2}} & \textcolor{myTableGray}{96.1} & \textcolor{myTableGray}{88.3} & \textbf{92.9}  & \textcolor{myTableGray}{94.7} & \textcolor{myTableGray}{71.8} & \textcolor{myTableGray}{\underline{98.9}} & \textcolor{myTableGray}{80.2} & \textcolor{myTableGray}{89.2} & 87.0  & \textbf{85.3} \\

\textbf{UniSim}$^{\clubsuit}$ LL-N-0.5B  & \textcolor{myTableGray}{89.8} & \textcolor{myTableGray}{70.0} & \textcolor{myTableGray}{85.3} & 81.7 & \textcolor{myTableGray}{69.2} & \textcolor{myTableGray}{80.7} & \textcolor{myTableGray}{66.7} & \textcolor{myTableGray}{90.8} & \textbf{\textcolor{myTableGray}{92.7}} & 80.0 & \textcolor{myTableGray}{75.4} & \textbf{\textcolor{myTableGray}{99.9}} & \underline{\textcolor{myTableGray}{89.2}} & 88.2  & \textcolor{myTableGray}{94.3} & \textcolor{myTableGray}{77.6} & \textcolor{myTableGray}{97.0} & \textcolor{myTableGray}{\underline{80.6}} & \textcolor{myTableGray}{89.8} & 87.9  & \underline{84.4} \\

 \textbf{UniSim}$^{\clubsuit, v_1}$ LL-N-0.5B  & \textcolor{myTableGray}{91.7} & \textcolor{myTableGray}{68.4} & \textcolor{myTableGray}{85.3} & 81.8 & \textcolor{myTableGray}{72.8} & \textcolor{myTableGray}{77.7} & \textcolor{myTableGray}{65.8} & \underline{96.2} & \textcolor{myTableGray}{91.2} & \textbf{80.7} & \textcolor{myTableGray}{74.4} & \underline{\textcolor{myTableGray}{99.8}} & \textcolor{myTableGray}{89.0} & 87.7  & \textcolor{myTableGray}{\underline{94.9}} & \textcolor{myTableGray}{70.3} & \textcolor{myTableGray}{97.7} & \textcolor{myTableGray}{79.6} & \textcolor{myTableGray}{89.7} & 86.4  & 84.2 \\

\textbf{UniSim}$^{\clubsuit, v_1}$ LL-N-7B & 92.7 & 67.6 & 86.6 & 82.3 & 60.2 & 72.6 & 65.2 & \textbf{97.7} & \underline{91.7} & 	77.5 & 71.2 & \textbf{99.9} & \textbf{90.7} & 87.3 & 93.4 & 73.9 & 96.6 & \textbf{81.2} & \underline{89.9} & 87.0 & 83.5 \\

\bottomrule
\end{tabular}}
\caption{\textbf{Full evaluation on the \textit{Core 2AFC Tasks} of \unisimbnc.} 
We complement the results of Table~\ref{tab:unisim-result} with additional metrics. 
LMM-based models are distinguished with the ${\clubsuit}$ symbol, while models highlighted with color are specialized in individual tasks (e.g., DS is specialized for the \imgtwoafc task). For LLaVA-based UniSim, $v_1$ is trained on perceptual data only (while the default version also uses the multi-image portion of LLaVA-NeXT data, see App.~\ref{sec:additional_experiments}).
The datasets used for training each model are indicated as superscripts next to their names. 
\textbf{Observations:} (1) Specialized models generally perform worse than general-purpose models on tasks outside their training domain, highlighting a significant lack of generalization. For example, the HPSv2 model, which is specialized for the \ittwoafc task, performs worse than the baseline (ViT-H/14) on the closely related \texttwoafc task. (2) UniSim ranks as the first or second best across nearly all tasks, demonstrating the feasibility of training a unified multi-modal metric capable of handling diverse and widely-used tasks.}
\label{tab:unisim-result-complete}
\end{table*}

\paragraph{\unisimlmm: LMM-based Perceptual Metric.}
To train \unisimlmm, we choose the LLaVA-NeXT~\cite{li2024llava} as the base model leveraging its advanced capability to handle multi-image inputs and image-text interleaved formats. LLaVA-NeXT, which relies on SigLIP-400M/14 vision encoder and the Qwen-1.5 language model (LLM), has two versions with different sizes: \href{https://huggingface.co/lmms-lab/llava-next-interleave-qwen-0.5b}{LLaVA-NeXT-0.5B} and \href{https://huggingface.co/lmms-lab/llava-next-interleave-qwen-7b}{LLaVA-NeXT-7B}.

One significant challenge in fine-tuning LMMs for perceptual tasks is that the ground truth typically consists of a single word representing the model's prediction between two alternatives. For training, we initially utilized our unified perceptual dataset, see Table~\ref{tab:unisim-train-data}, annotated with four distinct tasks: \nimgtwoafc (120K samples), \nittwoafc (300K samples), \ntexttwoafc (300K samples), and \niqa (120K samples). It is important to note that the number of samples for each task varies based on the complexity of the respective task. Additionally, we create another training dataset for \unisimlmm, which incorporates the multi-image section of the M4-Instruct dataset~\cite{li2024llava}, consisting of 500K samples, added to the UniSim data. We discuss in App.~\ref{sec:additional_experiments} how including this additional non-perceptual data for training helps improving generalization.
For the LLaVA-NeXT-0.5B the entire model, including the vision tower, adapter, and language model, is fine-tuned with learning rate $10 ^ {-5}$ for all components, except $2 \times 10^{-6}$ for the vision tower. While for LLaVA-NeXT-7B the adapter, and language model, are fine-tuned with $2 \times 10^{-6}$ learning rate to avoid overfitting to the training data. Weight decay is disabled, and a warm-up ratio of 0.03 of the total training steps is applied. The training is performed for a single epoch, following the standard practice for training LMMs.



\section{Additional Experiments} \label{sec:additional_experiments}
In this section, we begin by providing details on the evaluation setup. Next, we present the complete versions of Tables~\ref{tab:unisim-result} and ~\ref{tab:unisim-result-unseen}, including the detailed results over datasets and models omitted in the main part. Finally, we discuss the variations in the \unisimlmm models, focusing on differences in their size and training data.

\paragraph{Evaluation setup.}
While evaluating encoder-based perceptual metrics on \niqa, we test two approaches with the encoder models: first, a naive approach computes the alignment between the prompt \texttt{``A high quality photo.''} (i.e, the reference) and the two alternative images.
Second, we apply the CLIP-IQA technique from \cite{wang2023exploring}, where for each image one measures the similarity to two opposite prompts (\texttt{``Good photo.''}, \texttt{``Bad photo.''}), and obtains a quality score as the similarity to the first prompt after softmax normalization. The image with higher quality score is then chosen.
For each model we test both approaches and report the results of the one which performs best on average on the task.
Finally, we use the same two approaches for \npaa, again reporting the best-performing one for the task.
For evaluating the LMM-based models, we use specific instructions tailored to each perceptual task. These instructions are detailed in Table~\ref{tab:prompts}.

\begin{table*}[t]
\centering \small
\tabcolsep=1.2pt
\extrarowheight=1.5pt
\myindent=4mm
\newl=7mm
\newcommand{\myangle}{90}
 \scalebox{1.}{
\begin{tabular}{L{34mm}||
*{5}{C{\newl}}
>{\columncolor{myPurple!50}}C{\newl}|
*{2}{C{\newl}}
>{\columncolor{myOrange!50}}C{\newl}|
*{2}{C{\newl}}
>{\columncolor{myGray!70}}C{\newl}||
>{\columncolor{myGray!50}}C{\newl}
}
\toprule

& \multicolumn{6}{c}{\cellcolor{myPurple} \textbf{\paa}} & \multicolumn{3}{c}{\cellcolor{myOrange} \textbf{\ooo}} & \multicolumn{3}{c}{\cellcolor{myGray} \textbf{\ir}} & \textbf{Avg} \\
Models & 
\rotatebox{\myangle}{\sice-bri} &
\rotatebox{\myangle}{\koniq-bri} &
\rotatebox{\myangle}{\koniq-col} & 
\rotatebox{\myangle}{\koniq-con} & 
\rotatebox{\myangle}{\koniq-sha} & 
\rotatebox{\myangle}{average} &
\rotatebox{\myangle}{\imgnet} & 
\rotatebox{\myangle}{\cifarhooo} & 
 \rotatebox{\myangle}{average} &
\rotatebox{\myangle}{\oxford} &
\rotatebox{\myangle}{\paris} &
\rotatebox{\myangle}{average} &
\\
\toprule

\multicolumn{10}{l}{\textbf{General-purpose models}}\\
\toprule
\textbf{CLIP ViT-B/32}  & 97.1 & \underline{67.0} & 61.5 & 57.7 & 69.5 & 70.6 & \underline{68.2} & \textbf{74.3} & \textbf{71.3} & 28.1 & 59.6 & 43.8 &  61.9\\  
\textbf{CLIP ViT-L/14} & 94.5 & 58.5 & 57.6 & 58.2 & 65.4 &  66.8 & 69.4 &  62.1 & 65.8 & 31.8 & 59.3 & 45.5 & 59.4\\

\textbf{CLIP ViT-H/14} & 96.3 & 66.1 & 57.0 & \underline{61.1} & 60.5 & 68.2 & 66.7 & \underline{73.9} & 70.3 & \underline{36.8} & \underline{63.6} & \underline{50.2} & \underline{62.9}\\

\textbf{SigLIP 400m} & \underline{98.0} & 61.4 & 63.1  & 56.7 & 64.0 & 68.6 & 67.2 & 72.3 & 69.8 & \textbf{37.1} & \textbf{68.7} & \textbf{52.9} & \textbf{63.7}\\

\textbf{BLIP ViT-L/14} & 94.2 & 64.3 & 54.6 & 57.6 & 59.8 & 66.1 & 61.6 & 65.2 & 63.4 & 19.0 & 52.9 & 35.9 & 55.1\\ 

\textbf{LLaVA NeXT-0.5B$^{\clubsuit}$} & 87.6 & 57.8 & 
 63.4 & 54.5 & 56.3 & 63.9 & 34.3 & 33.0 & 33.7 & - & - & - & -\\
\textbf{LLaVA NeXT-7B$^{\clubsuit}$} &  92.7 & 66.2 & 64.1 & 58.2 & 60.4 & 68.3 & 56.5 & 65.5 & 61.0 & - & - & -& -\\
\textbf{Mantis Idefics-8b$^{\clubsuit}$} & 97.0 & 60.7 & 62.7 & 61.0 & 59.7 & 68.2 & 44.0 & 44.1 & 44.1 & - & - & -& -\\
\toprule

\multicolumn{10}{l}{\textbf{Specialized models}} \\
\toprule
\textbf{DS}$^{(1)}$ ViT-B/32 & \textbf{99.0} & 66.3 & 63.2 & 58.7 &	66.1 & 70.7 & 59.4 & 63.4 & 61.4  & 25.2 & 50.7 & 38.0 & 56.7 \\
\textbf{DS}$^{(1)}$ Ensemble & - & - & - & - & - & - & 64.8 & 69.1 & 67.0 & 27.3 & 57.2 & 42.2 & -\\
\textbf{IR}$^{(3)}$ ViT-L/14 & 91.4 & 62.2 & 57.0 & 56.2 &	58.8 & 65.1 & 67.6 &72.7 & 70.2 & 24.2 & 59.3 & 41.7 & 59.0\\
\textbf{HPSv2}$^{(4)}$ ViT-H/14 & 92.9 & 65.0 & 59.1 & \textbf{62.9} &	59.7 & 67.9 & 51.1 & 61.7 & 56.4 & 23.0 & 49.9 & 36.4 & 53.6\\
\textbf{PAC-S}$^{(5)}$ ViT-L/14 & 88.8 & 67.5 &	60.0 & 57.7 & 54.8 & 65.8 & \textbf{69.2} &73.1 & \underline{71.2} & 33.7 & 62.2 & 48.0 & 61.6\\
\textbf{C2S$^{\clubsuit}$$^{(6,7)}$}mOwl-2 & 63.5 & 62.7 & 51.1 & 57.5 & 71.4 & 61.2 & - &  - & - & - & - & - & -\\
\textbf{LIQE$^{(6,7)}$} ViT-B/32 & 92.8 & 68.0 & 58.2 & 60.0 & \textbf{75.9} & \underline{71.0} & 65.7 & 54.4 & 60.1 & 12.8 & 24.8 & 18.8 & 49.9 \\
\toprule

\multicolumn{10}{l}{\textbf{Our models}$^{(\dagger)}$} \\
\toprule
\textbf{UniSim} ViT-B/32 & 97.8 & \textbf{67.9} & \textbf{65.4} & 60.0 & \underline{73.2} &	\textbf{72.9} & 60.1 &63.6 & 61.9 & 20.0 & 48.4 & 34.2 & 56.3 \\
\textbf{UniSim} ViT-L/14 & 95.4 & 62.1 & 60.8 & 59.3 & 60.3 & 67.6 & 49.6 & 57.7 & 53.7 & 15.8 & 34.3 & 25.1 & 48.8\\

\textbf{UniSim}$^{\clubsuit}$ LL-N-0.5B & 73.0 & 62.3 & \underline{64.8} & 60.1 & 63.8 & 64.8 & 23.7 &24.6 & 24.2 & - & - & - & - \\

\textbf{UniSim}$^{\clubsuit, v_1}$ LL-N-0.5B & 68.7 & 62.4 & 61.9 & 61.0 & 63.1 & 63.4 & 15.9 & 16.4 & 16.2 & - & - & - & - \\

\textbf{UniSim}$^{\clubsuit, v_1}$ LL-N-7B & 71.5 & 56.4 & 58.1 & 51.8 & 61.3 & 59.8 & 24.7 & 15.5 & 20.1 & - & - & - & - \\
\bottomrule
\end{tabular}}
\caption{\textbf{Detailed evaluation on the \gentasks of \unisimbnc.} To complement the results of Table~\ref{tab:unisim-result-unseen}, report the performance of the various perceptual metrics on each dataset included in the \gentasks, together with the average performance over tasks.
Moreover, we include additional baselines as done in Table~\ref{tab:unisim-result-complete}.
}
\label{tab:unisim-result-unseen-complete}
\end{table*}

\paragraph{Complete evaluation.}

Table~\ref{tab:unisim-result-complete} presents a comprehensive evaluation of perceptual metrics using \unisimbnc. The table includes SigLIP 400m~\cite{zhai2023sigmoid}, a variation of CLIP where the softmax function is replaced with a sigmoid function. Additionally, it features DreamSim Ensemble, which integrates DINO~\cite{caron2021emerging}, OpenCLIP~\cite{cherti2023openclip}, and the CLIP model for enhanced performance.
Table~\ref{tab:unisim-result-unseen-complete} provides a detailed evaluation of each dataset within \gentasks, offering a comprehensive overview of the strengths and weaknesses of each model.

\paragraph{Other analyses.}

As previously mentioned, two versions of LLaVA-NeXT (0.5B and 7B) are used to train the \unisimlmm models. A comparison of these versions (trained only on UniSim training data) is presented in Tables~\ref{tab:unisim-result-complete} and ~\ref{tab:unisim-result-unseen-complete} (marked as $v_1$). Notably, \unisimlmm-7B exhibits clear signs of overfitting, performing worse than its baseline on the left-out datasets in \coretasks and on most datasets in \gentasks. In contrast, \unisimlmm-0.5B demonstrates better generalization.

Moreover, we see that \unisimlmm-0.5B, trained on both the UniSim training data and a subset of the LLaVA-NeXT data, achieves better generalization performance than \unisimlmm-0.5B$^{v_1}$, trained only on the UniSim data (see Table~\ref{tab:unisim-result-unseen-complete}).
We hypothesize that such additional data reduces overfitting to the specific 2AFC data structure.

\end{document}